\documentclass[journal]{IEEEtran}
\usepackage{latexsym}
\usepackage{graphicx}
\usepackage{amsmath}
\usepackage{amssymb}
\usepackage{epsfig}
\usepackage{subfigure}
\usepackage{mathrsfs}
\usepackage{setspace}
\usepackage{epstopdf}
\usepackage{cite}
\usepackage{color}
\usepackage{algorithm} 
\usepackage{algorithmic} 

\makeatletter

\newcommand{\Rmnum}[1]{\expandafter\@slowromancap\romannumeral #1@}
\makeatother
\ifCLASSINFOpdf
\else
\fi
\begin{document}
%

\title{Removing rain streaks by a linear model}

%
%
%

\author{Yinglong~Wang,~\IEEEmembership{Student Member,~IEEE},
        Shuaicheng~Liu,~\IEEEmembership{Member,~IEEE},
        and Bing~Zeng,~\IEEEmembership{Fellow,~IEEE}
\thanks{The authors are with Institute of Image Processing, University of Electronic Science and Technology of China, Chengdu, Sichuan 611731, China.}
\thanks{All correspondences to B. Zeng (e-mail: eezeng@uestc.edu.cn).}

}

%
%

\markboth{IEEE Transactions on Image Processing,~Vol.~XX, No.~XX, XX~2018}%
{Shell \MakeLowercase{\textit{et al.}}: Removing rain streaks by a linear model}
%



\maketitle

\begin{abstract}
Removing rain streaks from a single image continues to draw attentions today in outdoor vision systems. In this paper, we present an efficient method to remove rain streaks. First, the location map of rain pixels needs to be known as precisely as possible, to which we implement a relatively accurate detection of rain streaks by utilizing two characteristics of rain streaks.The key component of our method is to represent the intensity of each detected rain pixel using a linear model: $p=\alpha s + \beta$,
where $p$ is the observed intensity of a rain pixel and $s$ represents the intensity of the background (i.e., before rain-affected). To solve
$\alpha$ and $\beta$ for each detected rain pixel, we concentrate on a window centered around it and form an $L_2$-norm cost function by considering all detected rain pixels within the window, where the corresponding rain-removed intensity of each detected rain pixel is estimated by some neighboring non-rain pixels. By minimizing this cost function, we determine $\alpha$ and $\beta$ so as to construct the final rain-removed pixel intensity. Compared with several state-of-the-art works, our proposed method can remove rain streaks from a single color image much more efficiently - it offers not only a better visual quality but also a speed-up of several times to one degree of magnitude.
\end{abstract}


%
\IEEEpeerreviewmaketitle

\section{Introduction}

%
%
%
%

\IEEEPARstart{B}{ecause} of rain's high reflection to light, rain usually is imaged as bright streaks in an image and influences the visual quality of the image. Hence, removing rain streaks in images has been necessary for most photographers. Garg and Nayar revealed that the visuality of rain is strongly related to some camera parameters \cite{Garg_2005_ICCV}. Photographers can tune these parameters (e.g., exposure time and field depth) to constrain the captured rain streaks. However, this method can avoid rain streaks only to a small extent. In addition, the majority of rain images are obtained by outdoor vision systems where it is difficult to tune the camera's parameters timely.

Rain streaks change the information conveyed by the original image. Therefore, the effectiveness of many computer vision algorithms that are based on some small features would be degraded severely. Though a majority of tracking and recognition algorithms is implemented for videos, rain-removal in the key frames of a video turns to play an important role.

Due to the inevitability of rain weather and the wide deployment of vision systems in practice, rain removal has been an important problem in computer vision for a long time. The related research for rain can date back to 1948 when Marshall and Palmer analyzed the relationship between the distribution of rain and its size \cite{Marshall_1948_JM}. Then, Nayar \textit{et al.} studied the visual manifestations of different weather conditions, including rain and snow \cite{Nayar_1999_ICCV}. Because of the randomness of rain's location, accurate detection of rain is a very difficult task. The early research works were mainly concentrating on rain removal in videos \cite{Garg_2004_CVPR,Bossu_2011_CV}, thanks to the strong correlation among neighboring video frames. Later on, the research focus gradually shifts to the study of rain removal in a single (color) image \cite{Xu_2012_CIS,Kim_2013_ICIP,Ding_2015_MTA,Fu_2011_ASSP,Kang_2012_TIP,Chen_2014_CSVT,Huang_2012_ICME,Huang_2014_TM,Chen_2013_ICCV,Luo_2015_ICCV,Li_2016_CVPR}.

The recent methods for rain removal from a single image can be classified into four categories. The first category is simply filtering-based where a nonlocal mean filter or guided filter is often used \cite{Xu_2012_CIS,Kim_2013_ICIP,Ding_2015_MTA}. Due to the use of a filter simply, its implementation is very fast. However, it can hardly produce a satisfactory performance consistently - either the output image is left over with some rain streaks or quite a few image details are lost so that the output image becomes blurred. The second category builds models for rain streaks \cite{Chen_2013_ICCV,Luo_2015_ICCV,Li_2016_CVPR}. These models attempt to discriminate rain streaks from the background. However, it often happens that some details of the image will be mistreated as rain streaks. The third category, which seems more reasonable, is to form a 2-step processing \cite{Fu_2011_ASSP,Kang_2012_TIP,Chen_2014_CSVT}. Specifically, a low-pass filtering is first used to decompose a rain image into the low-frequency part and high-frequency part. While the low-frequency part can be made free of rain as much as possible, some descriptors can be applied on the high-frequency part to further extract the image details to be added back into the low-frequency part. The last category combines deep learning methods with the rain removal
task and obtains excellent results by designing appropriate deep networks\cite{Fu_2017_CVPR,Yang_2017_CVPR,Zhang_2017_arXiv,Fu_2017_TIP}.

\begin{figure*}[t]
\begin{center}
\begin{minipage}{0.16\linewidth}
\centering{\includegraphics[width=1\linewidth]{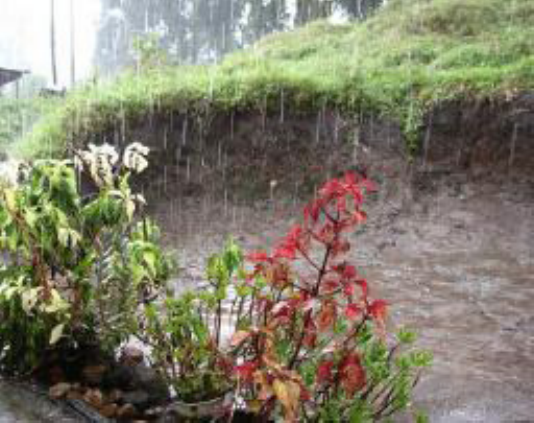}}
\centerline{(a)}
\end{minipage}
\hfill
\begin{minipage}{.16\linewidth}
\centering{\includegraphics[width=1\linewidth]{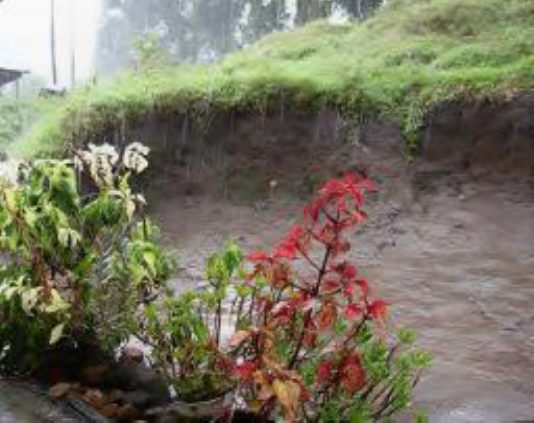}}
\centerline{(b)}
\end{minipage}
\hfill
\begin{minipage}{0.16\linewidth}
\centering{\includegraphics[width=1\linewidth]{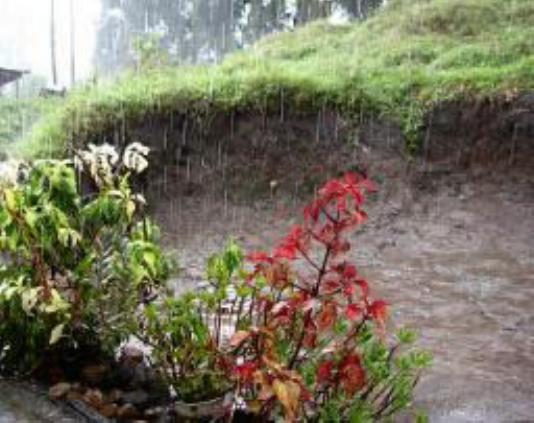}}
\centerline{(c)}
\end{minipage}
\hfill
\begin{minipage}{0.16\linewidth}
\centering{\includegraphics[width=1\linewidth]{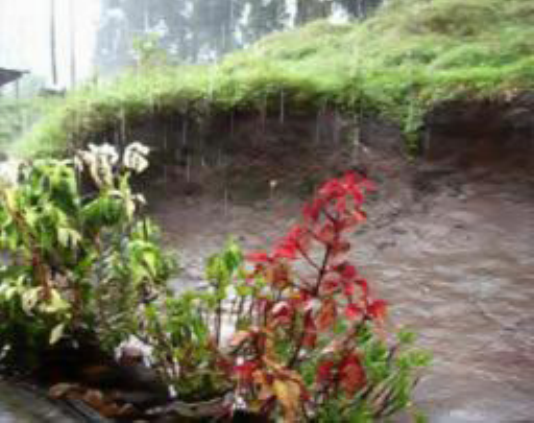}}
\centerline{(d)}
\end{minipage}
\hfill
\begin{minipage}{0.16\linewidth}
\centering{\includegraphics[width=1\linewidth]{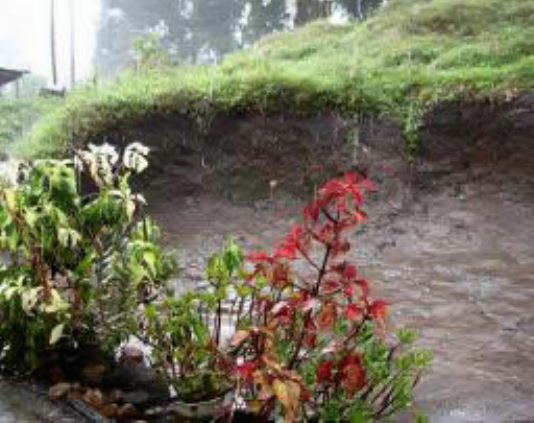}}
\centerline{(e)}
\end{minipage}
\hfill
\begin{minipage}{.16\linewidth}
\centering{\includegraphics[width=1\linewidth]{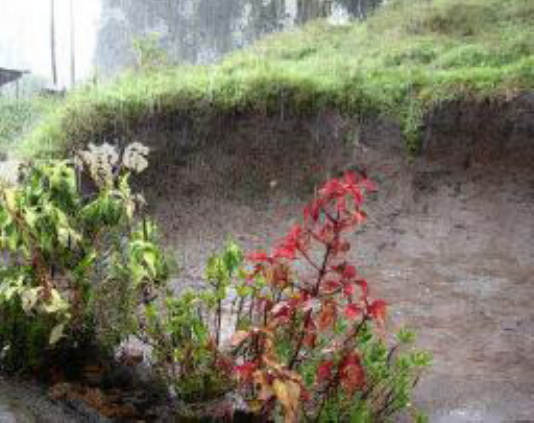}}
\centerline{(f)}
\end{minipage}
\end{center}
\caption{An example of the rain-removed results: (a) original rain image, (b) result of Li \textit{et al.} (in CVPR-2016) \cite{Li_2016_CVPR}, (c) result of Luo \textit{et al.} (in ICCV-2015) \cite{Luo_2015_ICCV}, (d) result of Zhang \textit{et al.} (in CVPR-2018) \cite{Zhang_2018_CVPR}, (e) result of Fu \textit{et al.} (in CVPR-2017) \cite{Fu_2017_CVPR}, (f) our result.}
\label{fig:test12}
\end{figure*}

Our formulation to remove rain streaks in a single color image also includes two steps. In the first step, we try to detect rain streaks by utilizing two characteristics of rain streaks. Then, a linear model is built to remove rain streaks in the second step.

\textbf{Step-1:} In order to remove rain streaks as much as possible, we hope that all rain streaks can be detected. However, the existing methods are difficult to obtain the locations of rain accurately and two kinds of detection errors usually occur:
\begin{enumerate}
   \item some rain streaks are missed, and
   \item some non-rain image details are mis-detected as rain.
\end{enumerate}

If rain streaks are missed in the detection, they will remain in the final result. Hence, we try our best to avoid this detection error in our work.
In our extensive tests, we found that our detection method capture nearly all rain streaks for a big majority of the test images. This is largely because that raindrops usually have strong reflection to light so that their pixel intensities are apparently larger as compared to the background pixels. Even when some rain streaks are missed, the final results could still be acceptable visually, because those missed rain-streaks have color components that are very similar to the background.

On the other hand, it is inevitable that some non-rain image details will be mis-detected as rain streaks. To reduce its influence, we try to revise the error detection by an eigen color method \cite{Tsai_2008_IET_CV}. To this end, two characteristics of rain streaks are used:
\begin{itemize}
\item rain usually possesses a higher reflection to light as compared to its neighboring non-rain objects, thus leading to larger intensities, and
\item rain is semi-transparent and colorless so as to present a gray color in the image.
\end{itemize}
These two characteristics are rather robust and have been utilized to rain removal works before, such as \cite{Chen_2014_CSVT} and \cite{Wang_2016_ICIP}.

\textbf{Step-2:} We follow the imaging principle of rain pixels to build a physical model to represent each rain pixel. In reality, there are many factors influencing the imaging of rain pixels, such as light, wind, and even the background. By a reasonable approximation, we simplify the imaging of rain into a linear function:
\begin{equation}\label{eq:linear_model}
p= \alpha s + \beta
\end{equation}
where $s$ is the pixel intensity of the scene before being affected by rain (which is unknown), $p$ is the observed intensity of rain pixel, $\alpha$ and $\beta$ are the parameters of the linear model, respectively. Our goal is to determine $\alpha$ and $\beta$ for each rain pixel so that $s$ can be reconstructed optimally.

\textbf{Advantages of our approach:} In our work, we propose to perform a relatively accurate detection of rain streaks to make sure that nearly no rain streaks are missed. Then, based on the linear model for rain pixels, we determine the involved parameters by optimizing a convex loss function. Since our proposed processing happens only on all detected rain pixels, other non-rain image details remain in the final result, which plays an important role in preserving image details. It can be seen from later sections that our algorithm produces higher PSNR/SSIM values for most rendered rain images compared with some state-of-the-art traditional methods. The optimization formulated in our work is a convex one. We can obtain the global optimal solution and avoid complex iterative calculation. Hence, our algorithm offers a speed-up of several times to more than one degree of magnitude when compared with several recent state-of-the-art works, thanks to the linear model and the detection of rain streaks developed in our work.

Furthermore, our linear model shows good robustness for removing rain streaks no matter on ordinary rain images or even on challenging heavy rain images. Another important point is that our algorithm is not a memory-consuming one. During experiment, we found that some algorithms (e.g., \cite{Luo_2015_ICCV}) have high requirements for computer memory. If the input rain images are relatively large, they will lead to memory overflow easily. By testing, our algorithm can deal with large rain images even on ordinarily configured computer. Through relatively complete comparison, our results prove to outperform those state-of-the-art works and be comparable to the deep learning based works both subjectively and objectively - referring to Fig. \ref{fig:test12} for one set of results.

The remainder of this paper is organized as follows. We briefly review the existing rain-removal methods in Section II. In Section III, we present the details of our rain detection method. The linear model for the imaging of rain pixels and the associated optimization to determine the involved parameter are described in Section IV. In Section V, we show the experiment results of our algorithm and make objective and subjective comparison with several state-of-the-art works. Finally, we conclude this paper in Section VI.

\section{Related Works}

\textbf{Rain removal from videos in the spatial domain:} Early work on detection and removal of rain is mainly focused on videos by making use of the correlation among video frames. In \cite{Garg_2004_CVPR}, Garg and Nayar analyzed the visual effect of rain on an imaging system. In order to detect and remove rain streaks from videos, they developed a correlation model to describe the dynamics of rain and a motion blur model to explain the photometry of rain. To make the study more complete, they  further revealed that the appearance of rain is the interaction result of the lighting direction, the viewing direction, and the oscillating shape of
a raindrop \cite{Garg_2006_TG}. Then, they built a new appearance model for rain (which is based on the raindrop's oscillation model) and rendered rain streaks. In \cite{Garg_2007_CV}, they further analyzed the visual effect of rain and the factors that influence it. In order to detect and remove rain streaks in videos, they developed a photometric model that describes the intensities caused by individual rain streaks and a dynamic model that reflects the spatio-temporal properties of rain.

Based on the temporal and chromatic characteristics of rain streaks in video, Zhang \textit{et al.} proposed another rain detection and removal algorithm in \cite{Zhang_2006_ICME}. This work shows that rain streaks do not always influence a certain area in videos. Besides, the intensity changes of three color components (namely, $\Delta R$, $\Delta G$, $\Delta B$) of a pixel are approximately equal to each other, i.e., $\Delta R \approx \Delta G \approx \Delta B$. By these two characteristics, rain streaks are detected and removed in videos. However, this method can only deal with videos that are captured by a stationary camera.

In \cite{Brewer_2008_CS}, Brewer and Liu utilized three characteristics of rain to detect rain streaks in video, and the detected rain streaks are removed by calculating the mean value of two neighbouring frames. In \cite{Bossu_2011_CV}, Bossu \textit{et al.} proposed the histogram of orientation of streaks (HOS) to detect rain streaks in image sequences.
Specifically, this method decomposes an image sequence into foreground and background by a Gaussian mixture model and rain streaks are separated into the foreground. Then, HOS which follows a model of Gaussian-uniform mixture is calculated to detect rain streaks more accurately.

\textbf{Rain removal from videos in the frequency domain:} In \cite{Barnum_2007_PACV}, Barnum \textit{et al.} detected rain streaks in the frequency domain. Specifically, they developed a physical model to simulate the general shape of rain and its brightness. Combined with the statistical properties of rain, this model is utilized to determine the influence of rain on the frequency of image sequences. Once the frequency of rain is detected, they will be constrained to obtain the rain-removed image sequences. Later on, in order to analyze the global effect of rain in the frequency space, they built a shape and appearance model for single rain streak in the image space \cite{Barnum_2010_CV}. Then, this model is combined with the statistical properties of rain to create another model to describe the global effect of rain in the frequency space.

\textbf{Single image rain removal:} In \cite{Roser_2009_CV}, monocular rain detection is implemented by Roser and Geiger in a single image, in which a photometric raindrop model is utilized. Meanwhile, Halimeh and Roser detected raindrops on the car windshield in a single image by utilizing the standard interesting point detector \cite{Halimeh_2009_VS}. In this algorithm, they built a model for the geometric shape of raindrops and studied the relationship between raindrops and the environment.

To the best of our knowledge, in \cite{Fu_2011_ASSP}, Fu \textit{et al.} accomplished the rain-removal task in single images for the first time by representing the image signal sparsely \cite{Aharon_2006_SP}. Kang \textit{et al.} \cite{Kang_2012_TIP} and Chen \textit{et al.} \cite{Chen_2014_CSVT} followed a similar framework and demonstrated some improved results. In particular, Kang \textit{et al.} identified the dictionary atoms \cite{Mairal_2010_MLR} of rain streaks by utilizing the histogram of oriented gradients (HOG) \cite{Dalal_2005_CVPR}, while Chen \textit{et al.} developed the depth of field (DoF) to extract more image details from some high-frequency components. Some other learning-based image decomposition methods were also proposed to remove rain in a single image \cite{Huang_2012_ICME,Huang_2014_TM}. They follow the similar formulation used in \cite{Fu_2011_ASSP,Kang_2012_TIP,Chen_2014_CSVT}. In \cite{Huang_2012_ICME}, a context-constrained image segmentation on the input image is implemented. In \cite{Huang_2014_TM}, a unsupervised clustering on the observed dictionary atoms is implemented via affinity propagation, which makes the image-dependent components with similar context information be identified.

Meanwhile, Xu \textit{et al.} \cite{Xu_2012_CIS} developed a rain-free guidance image and then utilized the guided filter \cite{He_2013_PAMI} to remove rain. Luo \textit{et al.} separated a rain image into the rain layer and de-rained layer by a nonlinear generative model (screen blend model) \cite{Luo_2015_ICCV}. Specifically, they approximated the rain and de-rained layers by high discriminative codes over a learned dictionary. Kim \textit{et al.} \cite{Kim_2013_ICIP} detected rain streaks in a single image by combining an elliptical shape model of rain and a kernel regression method \cite{Takeda_2007_TIP}. Then, they removed rain streaks by a non-local mean filter \cite{Buades_2005_CVPR}. Based on the fact that rain streaks usually reveal similar and repeated patterns in the image, Chen \textit{et al.} captured the spatio-temporally correlated rain streaks by a generalized low-rank model from matrix to tensor structure \cite{Chen_2013_ICCV}.

In \cite{Ding_2015_MTA}, Ding \textit{et al.} designed a guided $L_0$ smoothing filter and obtained a coarse rain-free image. The final rain-removed image is then acquired by a further minimization operation. Wang \textit{et al.} analyzed the characteristics of rain and proposed a rain-removal framework \cite{Wang_2016_ICIP}. Li \textit{et al.} utilized some patch-based priors of both the background and rain to separate a rain image into the rain layer and de-rained layer \cite{Li_2016_CVPR}. In \cite{Wang_2017_TIP}, we proposed a rain or snow removal algorithm based on a hierarchical approach. At first, a rain/snow image is decomposed into rain/snow-free low-frequency part and high-frequency by combining rain/snow detection and guided filter \cite{He_2013_PAMI}. Then, we extract 3-layers non-rain/snow details from the high-frequency part by utilizing sparse coding. Finally, we add the low-frequency part with the 3-layers image details together to obtain the rain/snow-removed image.

Recently, deep learning has been applied to the rain removal task. For instance, Fu \textit{et al.} extended ResNet to a deep detail network that reduces the input-to-output mapping range and makes the learning process easier \cite{Fu_2017_CVPR}, whereas the de-rained result is further improved by using some image-domain priori knowledge. They also built DerainNet to remove rain streaks \cite{Fu_2017_TIP} in which they use the high-frequency part of an image rather than the image itself during the training process. In the meantime, Zhang \textit{et al.} proposed a de-raining method called Image De-raining Conditional General Adversarial Network (ID-CGAN) \cite{Zhang_2017_arXiv}, which adds the quantitative, visual, and discriminative performance into the objective function and obtains good results. Yang \textit{et al.} designed a new rain image model and constructed a new deep learning architecture \cite{Yang_2017_CVPR}, which also achieves good rain-removed results. In a most recent work \cite{Zhang_2018_CVPR}, Zhang \textit{et al.} proposed a density-aware multi-stream densely connected convolutional neural network, called DID-MDN, to remove rain from single images. This network can automatically determine the rain-density information and then remove rain efficiently.

\section{Detection of Rain}
\label{sec:detection}

Because of the randomness of rain streaks in an image, the accurate detection of rain streaks is a challenging problem in the rain removal tasks,
even by deep learning based methods. Our goal here is to detect nearly all rain pixels in the input rain image and, at the same time, try to avoid the mis-detection of non-rain details as much as possible.

In \cite{Wang_2017_TIP}, we proposed a rough detection method of rain streaks based on the fact that the intensity of a rain pixel is usually larger than its neighboring non-rain pixels. In this work, we follow this approach but optimize it by utilizing the color characteristics of rain to reduce the mis-detection of non-rain details.

\begin{figure}[t]
\begin{center}
\begin{minipage}{0.24\linewidth}
\centering{\includegraphics[width=1\linewidth]{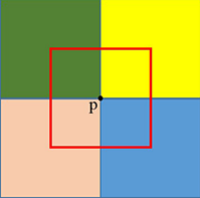}}
\end{minipage}
\end{center}
\caption{This figure shows five windows that are utilized to implement the rain detection in our work: $p$ is a pixel and located at the different location of five widows. Different colors denote four different windows with the pixel $p$ located at bottom-right (the green one), bottom-left (the yellow one), top-right (the orange one), and top-left (the blue one); the red square denotes the window with $p$ at the center location.}
\label{fig:five_windows}
\end{figure}

Let us use $I$ to denote an input rain image, which consists of three color channels as the input image is assumed to be a color one.
Because the intensity of a rain pixel is larger than its neighboring non-rain pixel in a small window, a pixel $p$ is considered as a rain pixel if its intensity is larger than the mean intensity of a window that includes $p$.
In this work, we modify slightly the detection rule as follow: $p$ is regarded as a rain pixel \emph{candidate} if the inequality
\begin{equation}\label{eq:detect_condition}
 p>\bar{p}^{(k)}+\mu,
\end{equation}
holds for all three color channels in five windows where $p$ locates at the center, top-left, top-right, bottom-left, and bottom-right position respectively, as shown in Fig. \ref{fig:five_windows}, and
\begin{equation}\label{eq:detect_condition1}
\bar{p}^{(k)}=\frac{\sum_{(m,n) \in \omega^{(k)}}p_{m,n}}{|\omega^{(k)}|}, k\!=\!\{1,2,3,4,5\},
\end{equation}
\noindent where $\omega^{(k)}$ represents a $7\times 7$ window.
The parameter $\mu$ is an empirical value which will be given later in the experimental part. For all pixels $p_{i,j}$ in image $I$, we implement the same detection. After the detection is completed, a binary map $B$ is generated in which $b_{i,j}$ is set to $0$ if a rain pixel candidate is detected at $(i,j)$.

As compared to \cite{Wang_2017_TIP}, a small incremental $\mu$ is added in (\ref{eq:detect_condition}). The reason is as follows: the intensity of some non-rain pixels could be smaller than the intensity of the neighboring rain pixel but larger than $\bar{p}^{k}$ so that $\mu$ can avoid mis-detecting this kind of non-rain pixels as rain.

By this detection method, nearly all rain pixels can be detected. The failure case is that in some rain images, rain pixels that have color components very similar to the background may be missed. Even some rain streaks are missed, the final results could still be acceptable visually, because those missed rain-streaks usually merge into the background.
At the same time, some non-rain details of the image are unfortunately mis-detected as rain pixels. An efficient way of identifying these mis-detections is described in the following.

Rain streaks usually possess a neutral color. According to this feature, Chen \textit{et al.} \cite{Chen_2014_CSVT} proposed to identify rain dictionary atoms by the eigen color feature \cite{Tsai_2008_IET_CV}. In our work, we utilize the color characteristics of rain to revise the mis-detected non-rain pixels. For a given pixel $p_{i,j}$, we use $[p^{R}_{i,j}, p^{G}_{i,j}, p^{B}_{i,j}]$ to represent its color vector in the RGB space. We transform this 3-D RGB space into a 2-D space as follows:

\begin{small}
\begin{equation}\label{eq:color_transform}
\begin{split}
 u_{i,j} & =\frac{2C_{i,j}-p^{G}_{i,j}-p^{B}_{i,j}} {C_{i,j}} \\
 v_{i,j} & =max \left \{
               \frac{C_{i,j}-p^{G}_{i,j}}{C_{i,j}}, \frac{C_{i,j}-p^{B}_{i,j}}{C_{i,j}}
            \right \}
\end{split}
\end{equation}
\end{small}

\noindent where $C_{i,j}=\frac{1}{3}(p^{R}_{i,j}+p^{G}_{i,j}+p^{B}_{i,j})$.

It is clear from (\ref{eq:color_transform}) that, after the transformation, any pixels having a neutral color will be clustered around $(0, 0)$ in the $u$-$v$ space. For each rain pixel candidate detected above, we transform their RGB values into the $u$-$v$ space to form a 2-D vector. If the magnitude of this 2-D vector (i.e.,the Euclidean distance to the origin of the $u$-$v$ space) is larger than a pre-set value $\epsilon$, $p_{i,j}$ is recognized as a mis-detected pixel. Consequently, we set its corresponding value in the location map $B$ back to $1$.

Now, the remaining content consists of real rain pixels and a few mis-detected non-rain details that are similar to rain in their size and eigen color. We cannot revise these remaining image details any more. From the statistics presented in Section 5, however, we will see that these mis-detected pixels contribute only a small percentage. By hundreds of tests on various kinds of rain images, their influence on the final rain-removed results is tolerable visually, we will analyze the reason later.

\section{A linear model and its optimization}
\label{sec:linear_model}

In this section, we build a linear model that is based on the imaging principle of a pixel to describe the influence of rain on the pixel intensity, followed by an optimization in which the involved parameters can be determined according to a convex loss function.

\subsection{The linear model}

When a scene is photographed, the intensity of each image pixel is  determined by an integral of the scene's irradiance over the entire exposure time $T$:
\begin{equation}\label{eq:irradiance_model0}
p_{s}= \int ^{T}_{0} R_{s} dt
\end{equation}
In a rainy scene, the intensity of a rain pixel $p$ can be expressed as:
\begin{equation}\label{eq:irradiance_model}
p= \int^{\tau}_{0} R_{r} dt + \int ^{T}_{\tau} R_{s} dt
\end{equation}
where $R_{r}$ and $R_{s}$ are the irradiance of raindrops and the scene, respectively, and $\tau$ is the occupation time of the raindrop during the entire exposure time $T$.

Raindrops have no regular shape. Hence, the irradiance $R_{r}$ of a raindrop is non-constant, and we utilize $\bar{R}_{r}$ to denote the time-averaged irradiance of raindrop. For a given pixel, the corresponding scene may change during the exposure time $T$ in video. However, in photography, the scene is stationary. Therefore, the irradiance $R_{s}$ of scene is constant. Then, (\ref{eq:irradiance_model}) can be rewritten as
\begin{equation}\label{eq:model_simplify}
p= \tau \bar{R_{r}} + (T-\tau) R_{s}
\end{equation}
Assuming that rain lasts for the entire exposure time, we have
\begin{equation}\label{eq:rain_intensity}
p_{r}= \int^{T}_{0} R_{r} dt = T \bar{R_{r}}
\end{equation}
For the scene, we make the same assumption and obtain
\begin{equation}\label{eq:scene_intensity}
s= \int^{T}_{0} R_{s} dt = T R_{s}
\end{equation}

By combining (\ref{eq:model_simplify}), (\ref{eq:rain_intensity}), and (\ref{eq:scene_intensity}) together, we obtain
\begin{equation}\label{eq:linear_model1}
p = \alpha s + \beta
\end{equation}
where $\alpha = (T - \tau) / T$ and $\beta=(1-\alpha) p_{r}$. This linear model establishes the relationship between the original scene intensity $s$ and the intensity $p$ after the scene is affected by rain.

As mentioned in \cite{Garg_2007_CV}, Gunn and Kinzer made an empirical study of the raindrop's terminal velocity in terms of the size of raindrop and found $v = 200 \sqrt{a}$, where $a$ is radius of a raindrop. In \cite{Garg_2007_CV}, Garg \textit{et al.} obtained a conservative upper bound of $\tau$ by simulating the process of rain's passing through the pin hole of a camera: $0 < \tau < 4a/v$. Then, the range of $\tau$ can be rewritten as $0 < \tau < \sqrt{a}/50$. Garg \textit{et al.} also found that the maximum value of $a$ is $3.5 \times 10^{-3}m$ so that the maximum value of $\tau$ is $1.2ms$ \cite{Garg_2007_CV}.

As analyzed above, the time $\tau$ is small compared with the entire exposure time $T$. Therefore, the values of $\alpha$ for all rain pixels in an image are nearly the same. Clearly, $\alpha$ is in the range $[0, 1]$. There are three situations according to the value of $\alpha$:
\begin{enumerate}[]
\item $\alpha = 1$ - the scene is not influenced by rain.
\item $\alpha \in (0, 1)$ - the most common situation where the pixel intensity follows the linear model described above.
\item $\alpha=0$ - implying that only rain is captured during the whole exposure time. In reality, this situation is impossible.
\end{enumerate}

When a raindrop falls down from the sky, its velocity becomes larger when getting closer to the ground. As we mentioned above, the maximum value of $\tau$ is $1.2ms$, which is usually small as compared with the exposure time of a camera $T$. This implies that $\alpha$ is close to $1$. We can further infer some imaging principles for rain from this model. For example, a low-intensity pixel would receive a larger intensity enhancement than a high-intensity pixel while being affected by rain. To prove this, the intensity enhancement is defined as:
\begin{equation}\label{eq:intensity_enhance}
\Delta s = p-s
\end{equation}
After applying our model, it will become:
\begin{equation}\label{eq:intensity_enhance1}
\Delta s = (\alpha -1)s + \beta
\end{equation}
Because $\alpha$ is less than $1$, the larger $s$ will lead to a smaller $\Delta s$.

Furthermore, our proposed linear model can be used to infer some other results obtained in previous works. For instance, in \cite{Zhang_2006_ICME}, Zhang \textit{et al.} demonstrated that the intensity changes of three color channels of a pixel are approximately equal to each other after being affected by rain. To prove this result, let us assume that $(R, G, B)$ is the color vector of a pixel and $(\hat{R}, \hat{G}, \hat{B})$ is the color vector after the pixel is affected by rain. Then, we have
\begin{equation}\label{eq:color_relation}
\begin{split}
&\hat{R} = \alpha R + \beta  \\
& \hat{G} = \alpha G + \beta  \\
& \hat{B} = \alpha B + \beta  \\
\end{split}
\end{equation}
the intensity change of each color channel is as follows:
\begin{equation}\label{eq:color_enhance1}
\begin{split}
&\Delta R = (\alpha -1)R + \beta  \\
& \Delta G = (\alpha -1)G + \beta  \\
& \Delta B = (\alpha -1)B + \beta  \\
\end{split}
\end{equation}
As we analyzed above, $\alpha$ is close to $1$.
Therefore, the intensity change of each color channel is approximately equal to each other, namely,
\begin{equation}\label{eq:color_enhance_equal}
\Delta R \approx \Delta G \approx \Delta B
\end{equation}

\begin{figure*}[!t]
\begin{center}
\begin{minipage}{0.24\linewidth}
\centering{\includegraphics[width=1\linewidth]{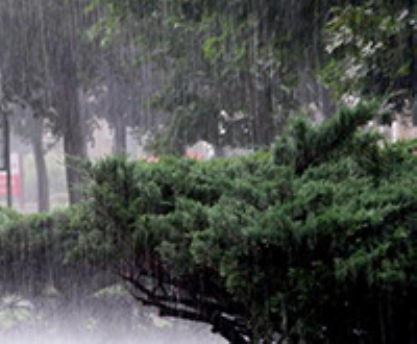}}
\centerline{(a)}
\end{minipage}
\begin{minipage}{0.24\linewidth}
\centering{\includegraphics[width=1\linewidth]{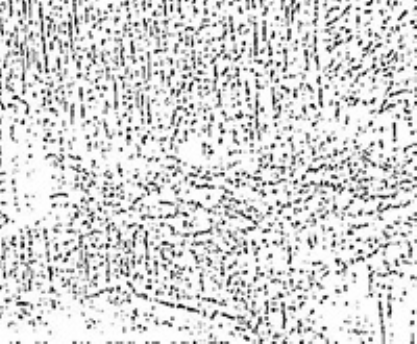}}
\centerline{(b)}
\end{minipage}
\hfill
\begin{minipage}{.24\linewidth}
\centering{\includegraphics[width=1\linewidth]{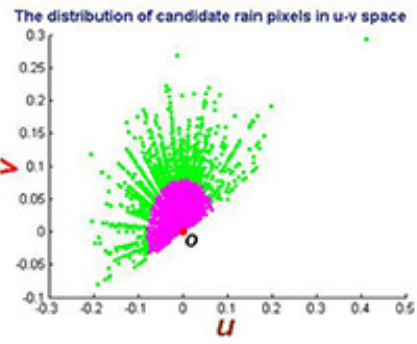}}
\centerline{(c)}
\end{minipage}
\hfill
\begin{minipage}{.24\linewidth}
\centering{\includegraphics[width=1\linewidth]{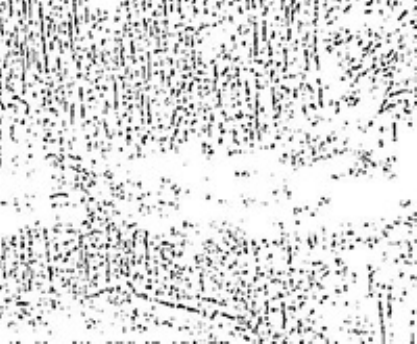}}
\centerline{(d)}
\end{minipage}
\end{center}
\caption{Step-by-step experimental results: (a) original rain image, (b) rough location map, (c) the distribution of detected candidate rain pixels in $u$-$v$ space, (d) revised location map.}
\label{fig:test55}
\end{figure*}

\subsection{Optimization}

In order to train the linear model for the given image $I$, we need to know the values of original pixels. Unfortunately, none of them is known in reality. The best we can do is to approximate each of them. In \cite{Brewer_2008_CS}, Brewer and Liu pointed out that a majority of rain streaks appears in the relatively constant areas in an image. This point can also be observed in rain images. Even in the image with complex background, there is still a small constant area around rain streaks. Otherwise, the rain streaks can not be seen. In such a relatively constant area, a weighted average intensity of the neighboring non-rain pixels around each rain pixel can be utilized to approximate its original intensity.
Such an approximation is supported by two reasons: (1) in the case of light rain, the intensity of the neighboring non-rain pixels is very close to the original intensity of the rain-affected pixel in a small window and (2) when encountered with heavy rain where the fog effect will appear, the weighted average serves as a de-hazing preprocessing and then the situation turns back to (1).

\textbf{Approximation of the original intensity of rain pixel.} \
For each rain pixel $p$, let us denote by $H=\{\textbf{\emph{h}}_k\}$ the set of color vectors of all non-rain pixels in the $N\times N$ window that is centered at $p$, i.e., $\textbf{\emph{h}}_k$, $k=1,2,\cdots,|H|$, represents the color vector of the $k$-th non-rain pixel in $H$,
and $\textbf{\emph{p}}$ is the observed color vector of rain pixel $p$.
To compute a weighted average of these non-rain pixels, the involved weights are calculated as follows:
\begin{equation}\label{eq:color_weight}
w_{k}=exp(- \frac{\| \textbf{\emph{h}}_{k} - \textbf{\emph{p}} \|_2^2}{\sigma^2}).
\end{equation}

For the rain pixel $p$, the approximation of its original color vector is
\begin{equation}\label{eq:approximation}
\textbf{\emph{q}} = \frac{\sum_{k \in H} w_{k}^{2} \textbf{\emph{h}}_{k}}{\sum_{k \in H} w_{k}^{2}}.
\end{equation}
We apply (\ref{eq:approximation}) to all detected rain pixels to obtain the approximations of their original intensities.

\begin{figure*}[t]
\begin{center}
\begin{minipage}{0.16\linewidth}
\centering{\includegraphics[width=1\linewidth]{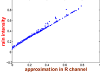}}
\centerline{(a)}
\end{minipage}
\hfill
\begin{minipage}{.16\linewidth}
\centering{\includegraphics[width=1\linewidth]{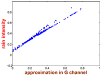}}
\centerline{(b)}
\end{minipage}
\hfill
\begin{minipage}{0.16\linewidth}
\centering{\includegraphics[width=1\linewidth]{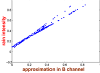}}
\centerline{(c)}
\end{minipage}
\hfill
\begin{minipage}{0.16\linewidth}
\centering{\includegraphics[width=1\linewidth]{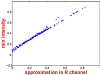}}
\centerline{(a1)}
\end{minipage}
\hfill
\begin{minipage}{.16\linewidth}
\centering{\includegraphics[width=1\linewidth]{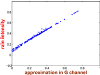}}
\centerline{(b1)}
\end{minipage}
\hfill
\begin{minipage}{0.16\linewidth}
\centering{\includegraphics[width=1\linewidth]{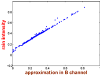}}
\centerline{(c1)}
\end{minipage}
\end{center}
\caption{A linear relationship between the intensities of rain pixels
and the approximations of their original intensities can be seen clearly in three color channels: (a) R channel, (b) G channel, and (c) B channel. Then, we manually revise the mis-detected non-rain pixels and reduce the influence of non-rain pixels on the linear model: (a1) R channel, (b1) G channel, and (c1) B channel.}
\label{fig:test55_statistics}
\end{figure*}

\textbf{Train the parameters of our linear model.} \ As analyzed above, the time $\tau$ is small compared with the entire exposure time $T$ and the parameters $\alpha$ and $\beta$ for all rain pixels in an image are approximately the same. Hence, we train the parameters of our model in a relatively large window, but only apply the resulted parameters to the rain pixel that is located at the window center.

Suppose that $p$ is a detected rain pixel, in one color channel $X$ (i.e., $X \in \{R, G, B\}$), we form a relatively large $M \times M$ window $\Omega$ centered at $p$ and all detected rain pixels $D=\{d_k\}$ in the window will be utilized. Let $Q=\{q_{k}\}$ as obtained by (\ref{eq:approximation}) be the approximations of the original intensities of rain pixels in $\Omega$ and $K = |D| = |Q|$.
According to our linear model (\ref{eq:linear_model1}), we have:
\begin{equation}\label{eq:linear_model_in_window}
d_{k}= \alpha q_{k} + \beta, \quad k=1, 2, ..., K.
\end{equation}

In order to determine $\{\alpha, \beta\}$, we minimize a loss function defined below:
\begin{equation}\label{eq:loss_function}
\begin{split}
 E(\alpha, \beta) =
 \sum^{K}_{k=1}
((d_{k} - \alpha q_{k} - \beta)^{2} + \lambda \alpha^{2})
\end{split}
\end{equation}
where $\lambda$ is the regularization parameter. This is a convex loss function and the solution is as follows:
\begin{equation}\label{eq:alpha}
\begin{split}
\alpha = \frac {\frac{1} {K} \sum^{K}_{k=1} d_{k}q_{k} - \bar{d} \bar{q}}
{\frac{1} {K} \sum^{K}_{k=1} q_{k}^{2} - \bar{q}^{2} + \lambda}
\end{split}
\end{equation}
\begin{equation}\label{eq:beta}
\begin{split}
\beta = \bar{d} - \alpha\bar{q}
\end{split}
\end{equation}
where $\bar{d} = \frac{1}{K} \sum^{K}_{k=1} d_{k}$,
$\bar{q}= \frac{1}{K} \sum^{K}_{k=1} q_{k}$, and $\frac{1} {K} \sum^{K}_{k=1} q_{k}^{2} - \bar{q}^{2}$ is the variance of $q_{k}, (k=1, 2, ..., K)$ in the window $\Omega$.

After obtaining $\alpha$ and $\beta$ for all detected rain pixels $D=\{d_k\}$, $k=1, 2, ..., K$ in the window $\Omega$, we only apply the model's parameters $\alpha$ and $\beta$ to the rain pixel $p$ that is the center of $\Omega$, so that its rain-removed intensity can be obtained as follows:
\begin{equation}\label{eq:rain_removed_value}
\begin{split}
s = (p-\beta) / \alpha.
\end{split}
\end{equation}

We implement the same processing on all detected rain pixels in color channel $X, X\in \{ R, G, B \}$ so that the rain-removed image $S$ can be obtained.

In order to make our work more clear, we summarize our algorithm in Algorithm \ref{alg:whole_algorithm}, whereas the values of the parameters used in our algorithm will be shown in the next section.

\begin{algorithm}
\renewcommand{\algorithmicrequire}{\textbf{Input:}}
\caption{Removing rain streaks by a linear model}
\label{alg:whole_algorithm}
\begin{algorithmic}
\REQUIRE input rain image $I$
\renewcommand{\algorithmicrequire}{\textbf{Step-1}}
\REQUIRE
\STATE obtain location $B$ using (\ref{eq:detect_condition})
\STATE transform the RGB space to the $u$-$v$ space by (\ref{eq:detect_condition1})
\STATE revise $B$ in the $u$-$v$ space by $\epsilon$
\renewcommand{\algorithmicrequire}{\textbf{Step-2}}
\REQUIRE
\FOR{$X \in \{ R,G,B \}$ }
\FOR{each detected rain pixel $p$ }
\STATE train $\{ \alpha, \beta \}$
by (\ref{eq:alpha}) and (\ref{eq:beta})
\STATE calculate the rain removed intensity $s$  by (\ref{eq:rain_removed_value})
\ENDFOR
\ENDFOR
\renewcommand{\algorithmicensure}{\textbf{Output:}}
\ENSURE rain-removed image $S$
\end{algorithmic}
\end{algorithm}

\section{Experimental results}

In this section we apply our algorithm to rain images. We first show some detailed intermediate experimental results and analysis for one test image.
Then, several state-of-the-art methods are selected to perform subjective and objective comparisons.

\subsection{Detailed experimental results}

We use the rain image in Fig. \ref{fig:test55}(a) as an example to run our algorithm and show the implementation details of our algorithm.

\textbf{Step-1.} We first detect the rain streaks by (\ref{eq:detect_condition}). The resulted binary location map is shown in Fig. \ref{fig:test55}(b) in which the black areas stand for the detected rain pixels.
In order to revise the mis-detected image details, we transform the RGB color space of each detected rain pixel candidate in $I$ into the $u$-$v$ space by (\ref{eq:detect_condition1}) and calculate its magnitude. The distribution of detected rain pixels in $u$-$v$ space are shown in Fig. \ref{fig:test55}(c). Then, we revise all rain pixel candidates whose magnitudes in the $u$-$v$ space are larger than the preset $\epsilon$ (the green part in Fig. \ref{fig:test55}(c)). The revised result is shown in Fig. \ref{fig:test55}(d), from which one can see that many rain pixel candidates have been revised.

\begin{figure*}[t]
\begin{center}
\begin{minipage}{0.24\linewidth}
\centering{\includegraphics[width=1\linewidth]{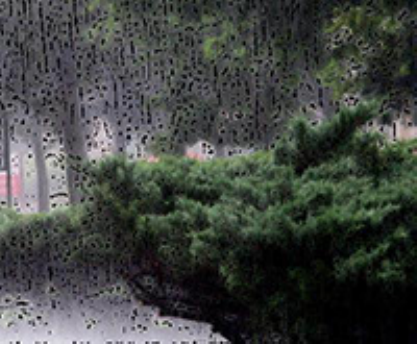}}
\centerline{(a)}
\end{minipage}
\hfill
\begin{minipage}{0.24\linewidth}
\centering{\includegraphics[width=1\linewidth]{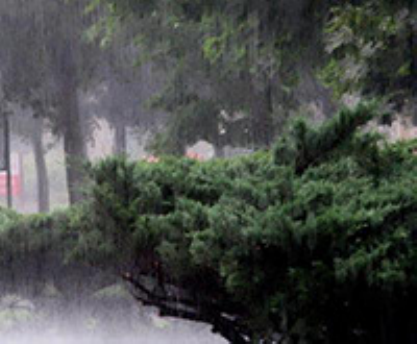}}
\centerline{(b)}
\end{minipage}
\hfill
\begin{minipage}{.24\linewidth}
\centering{\includegraphics[width=1\linewidth]{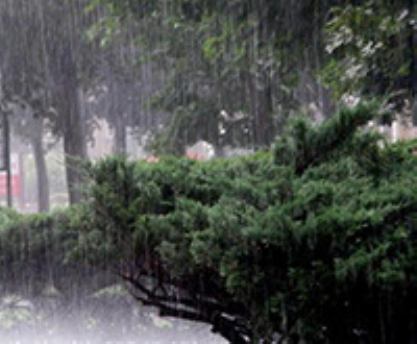}}
\centerline{(c)}
\end{minipage}
\begin{minipage}{.24\linewidth}
\centering{\includegraphics[width=1\linewidth]{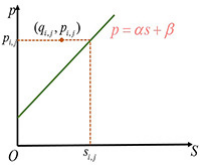}}
\centerline{(d)}
\end{minipage}
\end{center}
   \caption{(a) the result of multiplying binary location map with original rain image; (b) rain-removed result by our linear model; (c) the result of adding rain by the linear model on (b); (d) simulated diagram of the value of mis-detected non-rain details in rain-removed result.}
\label{fig:test55_add_rain}
\end{figure*}

\begin{table} [t]
\centering
\caption{The average time consumed by different methods on $256 \times 256$ color images.}
\begin{tabular}{lccccc}
\hline
Method     & \cite{Chen_2014_CSVT}   & \cite{Luo_2015_ICCV}  & \cite{Li_2016_CVPR}   & Ours   \\
Time(s)  &110.1  &75.7  &1257.4  & 7.68    \\
\hline
\end{tabular}
\label{tab:time}
\end{table}

\begin{table*} [t]
\small
\newcommand{\tabincell}[2]{\begin{tabular}{@{}#1@{}}#2\end{tabular}}
\centering
\caption{Rain image performances (top: \textbf{PSNR}, bottom: \textbf{SSIM}) of different methods (rows) on $10$ synthesized rain images (columns) against groundtruth.}
\begin{tabular}{l|c|c|c|c|c|c|c|c|c|c}
\hline
        & Image 1                       & Image 2                       & Image 3                       & Image 4                       & Image 5                       & Image 6                       & Image 7                       & Image 8                       & Image 9                       & Image 10                      \\
\hline
\cite{Chen_2014_CSVT} & \tabincell{c}{32.96 \\ 0.738} & \tabincell{c}{34.95 \\ 0.774} & \tabincell{c}{38.17 \\ 0.776} & \tabincell{c}{31.83 \\ 0.602} & \tabincell{c}{32.10 \\ 0.704} & \tabincell{c}{\textbf{34.14} \\ 0.784} & \tabincell{c}{34.37 \\ 0.788} & \tabincell{c}{\textbf{35.41} \\ 0.755} & \tabincell{c}{35.11 \\ 0.725} & \tabincell{c}{34.83 \\ 0.816}  \\
\hline
\cite{Luo_2015_ICCV}   & \tabincell{c}{32.42 \\ 0.820} & \tabincell{c}{33.71 \\ 0.838} & \tabincell{c}{35.92 \\ 0.784} & \tabincell{c}{29.44 \\ 0.790} & \tabincell{c}{30.41 \\ 0.878} & \tabincell{c}{32.96 \\ \textbf{0.843}} & \tabincell{c}{27.51 \\ 0.864} & \tabincell{c}{32.07 \\ 0.797} & \tabincell{c}{31.31 \\ 0.776} & \tabincell{c}{31.44 \\ 0.865}  \\
\hline
\cite{Li_2016_CVPR} & \tabincell{c}{32.96 \\ 0.691} & \tabincell{c}{33.03 \\ 0.746} & \tabincell{c}{36.39 \\ 0.682} & \tabincell{c}{29.52 \\ 0.669} & \tabincell{c}{30.31 \\ 0.686} & \tabincell{c}{32.36 \\ 0.747} & \tabincell{c}{31.21 \\ 0.706} & \tabincell{c}{33.49 \\ 0.720} & \tabincell{c}{29.90 \\ 0.697} & \tabincell{c}{31.83 \\ 0.734} \\
\hline
Ours   & \tabincell{c}{\textbf{33.77} \\ \textbf{0.869}} & \tabincell{c}{\textbf{39.34} \\ \textbf{0.847}} & \tabincell{c}{\textbf{38.23} \\ \textbf{0.794}} & \tabincell{c}{\textbf{33.33} \\ \textbf{0.796}} & \tabincell{c}{\textbf{34.86} \\ \textbf{0.896}} & \tabincell{c}{33.38 \\ 0.827} & \tabincell{c}{\textbf{34.89} \\ \textbf{0.871}} & \tabincell{c}{34.55 \\ \textbf{0.813}} & \tabincell{c}{\textbf{35.92} \\ \textbf{0.788}} & \tabincell{c}{\textbf{36.68} \\ \textbf{0.882}} \\
\hline
\end{tabular}
\label{tab:psnrssim_rain}
\end{table*}

\begin{table} [t]
\centering
\caption{User study result. The numbers are the percentages of votes which are obtained by each method.}
\begin{tabular}{lccccc}
\hline
Method     & \cite{Chen_2014_CSVT}   & \cite{Luo_2015_ICCV}  & \cite{Li_2016_CVPR}   & Ours   \\
Percentage  &10.54 \% &12.16 \% &7.50 \% & 69.80 \%    \\
\hline
\end{tabular}
\label{tab:statistics}
\end{table}

\begin{figure*}[t]
\begin{center}
\begin{minipage}{0.115\linewidth}
\centering{\includegraphics[width=1\linewidth]{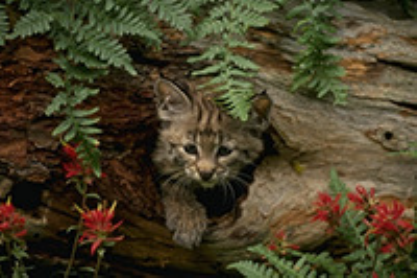}}
\end{minipage}
\hfill
\begin{minipage}{0.115\linewidth}
\centering{\includegraphics[width=1\linewidth]{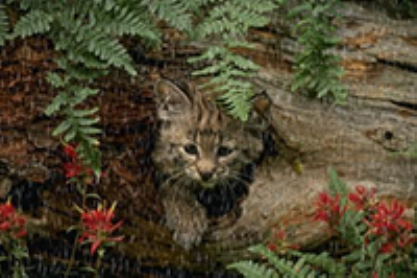}}
\end{minipage}
\hfill
\begin{minipage}{0.12\linewidth}
\centering{\includegraphics[width=1\linewidth]{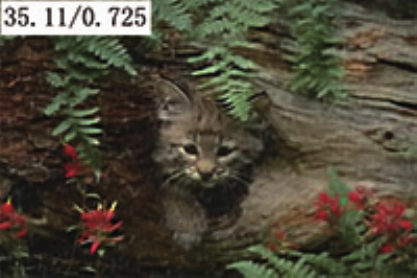}}
\end{minipage}
\hfill
\begin{minipage}{.115\linewidth}
\centering{\includegraphics[width=1\linewidth]{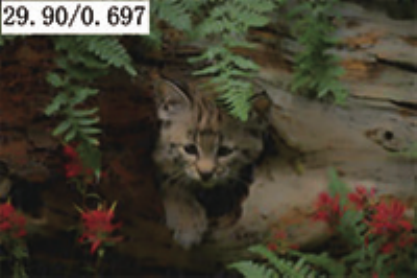}}
\end{minipage}
\hfill
\begin{minipage}{0.115\linewidth}
\centering{\includegraphics[width=1\linewidth]{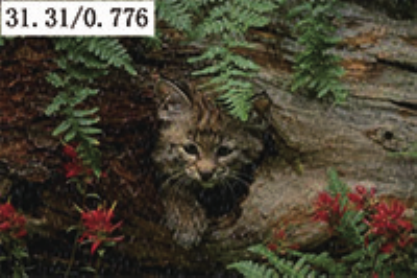}}
\end{minipage}
\hfill
\begin{minipage}{0.115\linewidth}
\centering{\includegraphics[width=1\linewidth]{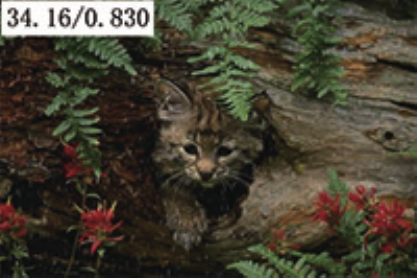}}
\end{minipage}
\hfill
\begin{minipage}{0.115\linewidth}
\centering{\includegraphics[width=1\linewidth]{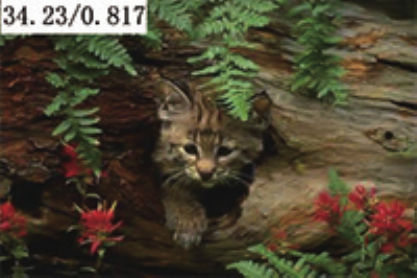}}
\end{minipage}
\hfill
\begin{minipage}{.115\linewidth}
\centering{\includegraphics[width=1\linewidth]{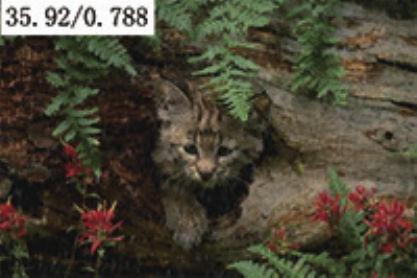}}
\end{minipage} \\ \vspace{0.5mm}
\vfill
\begin{minipage}{0.115\linewidth}
\centering{\includegraphics[width=1\linewidth]{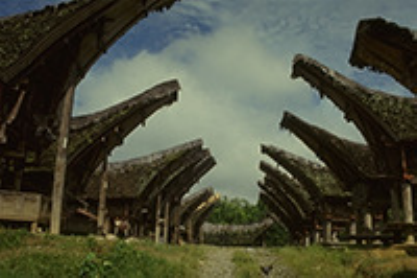}}
\centerline{(a)}
\end{minipage}
\hfill
\begin{minipage}{0.115\linewidth}
\centering{\includegraphics[width=1\linewidth]{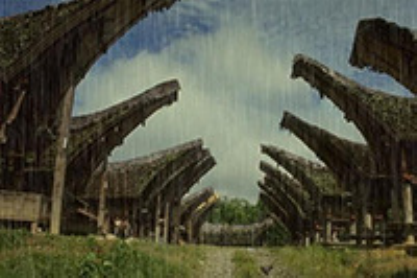}}
\centerline{(b)}
\end{minipage}
\hfill
\begin{minipage}{0.115\linewidth}
\centering{\includegraphics[width=1\linewidth]{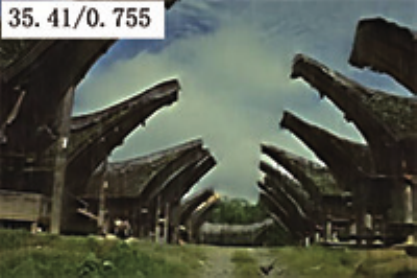}}
\centerline{(c)}
\end{minipage}
\hfill
\begin{minipage}{.115\linewidth}
\centering{\includegraphics[width=1\linewidth]{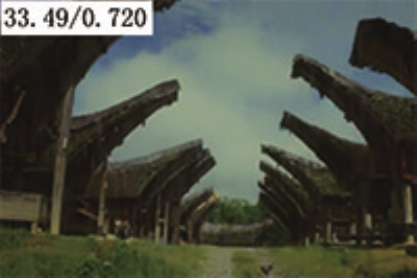}}
\centerline{(d)}
\end{minipage}
\hfill
\begin{minipage}{0.115\linewidth}
\centering{\includegraphics[width=1\linewidth]{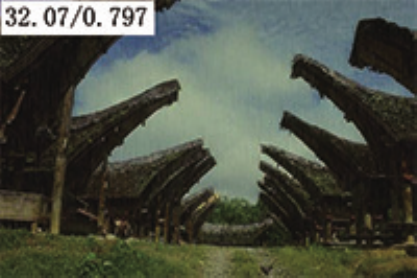}}
\centerline{(e)}
\end{minipage}
\hfill
\begin{minipage}{0.115\linewidth}
\centering{\includegraphics[width=1\linewidth]{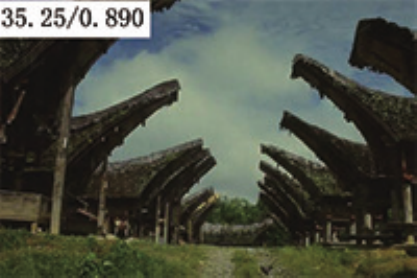}}
\centerline{(f)}
\end{minipage}
\hfill
\begin{minipage}{0.115\linewidth}
\centering{\includegraphics[width=1\linewidth]{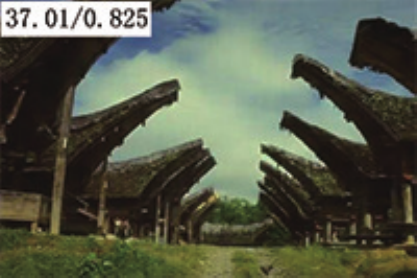}}
\centerline{(g)}
\end{minipage}
\hfill
\begin{minipage}{.115\linewidth}
\centering{\includegraphics[width=1\linewidth]{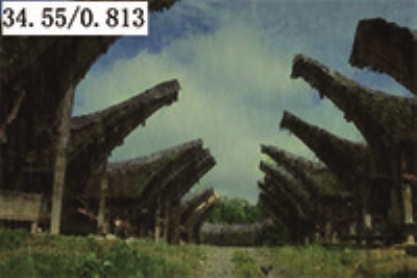}}
\centerline{(h)}
\end{minipage} \\ \vspace{0.5mm}
\end{center}
\caption{Some rain-removed results for synthetic rain images: (a) ground-truthes, (b) synthetic rain images, (c) results by Chen \textit{et al.} \cite{Chen_2014_CSVT}; (d) results of Li \textit{et al.} \cite{Li_2016_CVPR}, (e) results of Luo \textit{et al.} \cite{Luo_2015_ICCV}, (f) results of Fu \textit{et al.} \cite{Fu_2017_CVPR}, (g) results of Zhang \textit{et al.} \cite{Zhang_2018_CVPR} and (h) our results. We put PSNR/SSIM values of different methods at the top left corner of the images to facilitate comparisons.}
\label{fig:render_rain}
\end{figure*}

\begin{figure}[t]
\begin{center}
\begin{minipage}{.48\linewidth}
\centering{\includegraphics[width=1\linewidth]{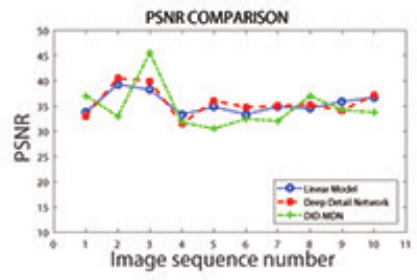}}
\centerline{(a)}
\end{minipage}
\begin{minipage}{.48\linewidth}
\centering{\includegraphics[width=1\linewidth]{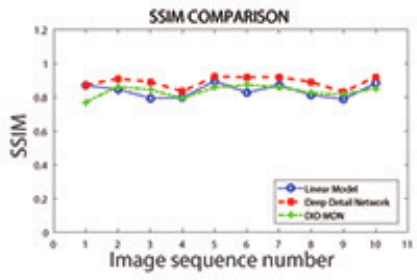}}
\centerline{(b)}
\end{minipage}
\end{center}
   \caption{(a) The PSNR values of three methods. (b) The SSIM values of three methods.}
\label{fig:psnr_ssim_graph_Fu}
\end{figure}

\textbf{Step-2.} For all detected rain pixels (i.e., those after the revising in Step-1), we calculate the approximations of their original intensities by (\ref{eq:approximation}). We plot in Fig. \ref{fig:test55_statistics}(a)-(c) the corresponding relationship between them, where 300 rain pixels are selected randomly in each of the R, G, and B channels. It can be seen from Fig. \ref{fig:test55_statistics} that the observed intensities of rain pixels and the approximations of their original intensities (before being affected by rain) indeed construct a good linear relationship in each of the three color channels.

In the meantime, we find that a few points deviate from the linear line in Fig. \ref{fig:test55_statistics}(a)-(c) and they are the mis-detected non-rain pixels. In order to verify this, we select true rain pixels manually to implement the same statistics and the results are shown in Fig. \ref{fig:test55_statistics}(a1)-(c1). We can see that most isolated points have been eliminated. Because those isolated points contribute only a small percentage, their influence on the training of the linear models is in the tolerable range.
To visually verify our detection method further, we use the binary location map to multiply the original rain image and the result is shown in Fig. \ref{fig:test55_add_rain}(a). It can be seen that no rain streaks are left. We would like to point out that very similar results have also been found in many other rain images.

By optimization in Algorithm \ref{alg:whole_algorithm}, we train the linear model and obtain the rain-removed result as shown in Fig. \ref{fig:test55_add_rain}(b). In order to verify the linear model further, we utilize the trained linear model to add rain streaks on the rain-removed image in Fig. \ref{fig:test55_add_rain}(b) and obtain the image in Fig. \ref{fig:test55_add_rain}(c). We can see that this calculated rain image is very similar to the original rain image in Fig. \ref{fig:test55}(a). In this particular example, we would like to report that the ranges of $\alpha$ and $\beta$ are $[0.91, 0.98]$ and $[0.10, 0.18]$ respectively. For other rain images, we can obtain the similar ranges of $\alpha$ and $\beta$, which implies a good consistency with our previous analysis, namely, the values of $\alpha$ are close to 1 and vary within a small range.

In our work, the parameters $\mu$, $\epsilon$, $\sigma$, $N$, $M$ and $\lambda$ are set as $0.01$, $0.08$, $9$, $13$, $85$, and $0.01^2$, respectively. Note that the input rain image has been normalized into the range $[0, 1]$.

Finally, we present an analysis of the intensities of mis-detected non-rain
details in the rain-removed results. The simulated diagram is shown in Fig. \ref{fig:test55_add_rain}(d), where $p_{i,j}$ denotes the observed intensity of a mis-detected non-rain pixel. According to the statistics of Fig. \ref{fig:test55_statistics}, mis-detected non-rain pixels often have high intensities and therefore are located above the linear line drawn by the trained linear model. When $\alpha$ of the model is close to $1$, $s_{i,j}$ would be lower than $p_{i,j}$. Hence, the intensities of mis-detected non-rain pixels will usually be reduced a little bit in the rain-removed results.
As we analyzed previously, non-rain pixels that have high intensities and whose color channels are nearly equal to each other will be mis-detected. After applying our linear model, their color channels are still approximately equal to each other, except for the slightly reduced intensities. This leads to a similar color appearance in the final result so that the image details are maintained well even mis-detection exists.

\begin{figure*}[t]
\begin{center}
\begin{minipage}{0.135\linewidth}
\centering{\includegraphics[width=1\linewidth]{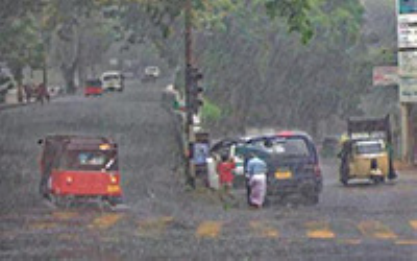}}
\end{minipage}
\hfill
\begin{minipage}{0.135\linewidth}
\centering{\includegraphics[width=1\linewidth]{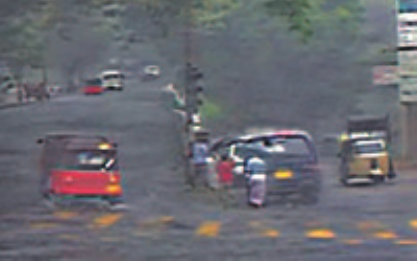}}
\end{minipage}
\hfill
\begin{minipage}{.135\linewidth}
\centering{\includegraphics[width=1\linewidth]{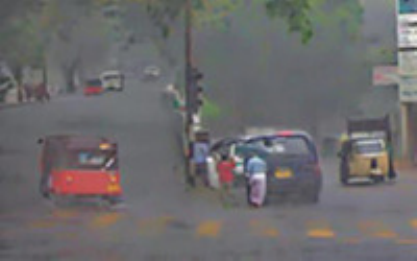}}
\end{minipage}
\hfill
\begin{minipage}{0.135\linewidth}
\centering{\includegraphics[width=1\linewidth]{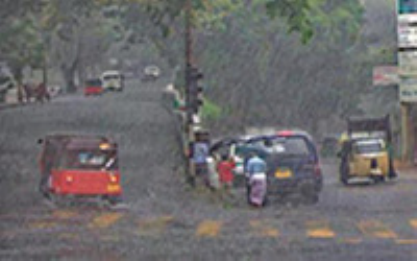}}
\end{minipage}
\hfill
\begin{minipage}{0.135\linewidth}
\centering{\includegraphics[width=1\linewidth]{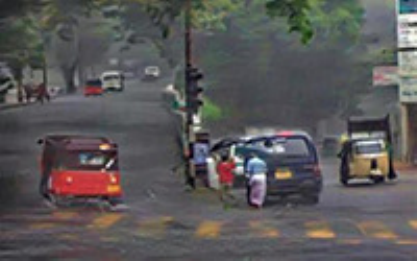}}
\end{minipage}
\hfill
\begin{minipage}{0.135\linewidth}
\centering{\includegraphics[width=1\linewidth]{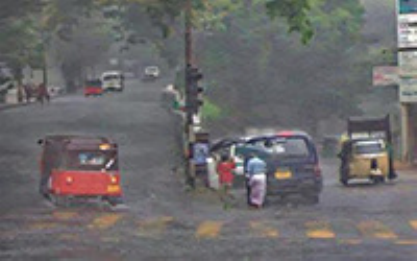}}
\end{minipage}
\hfill
\begin{minipage}{.135\linewidth}
\centering{\includegraphics[width=1\linewidth]{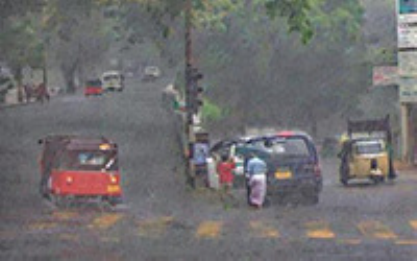}}
\end{minipage} \\ \vspace{0.5mm}
\vfill
\begin{minipage}{0.135\linewidth}
\centering{\includegraphics[width=1\linewidth]{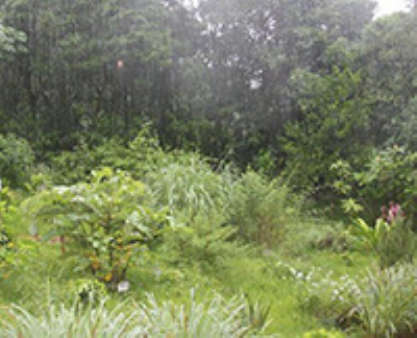}}
\end{minipage}
\hfill
\begin{minipage}{0.135\linewidth}
\centering{\includegraphics[width=1\linewidth]{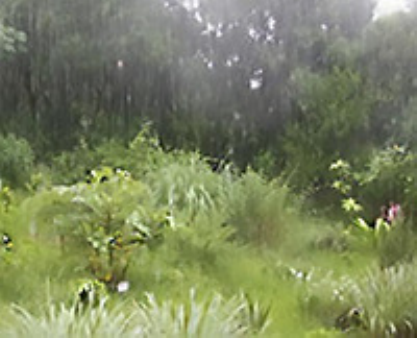}}
\end{minipage}
\hfill
\begin{minipage}{.135\linewidth}
\centering{\includegraphics[width=1\linewidth]{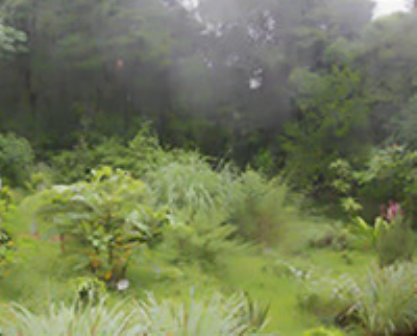}}
\end{minipage}
\hfill
\begin{minipage}{0.135\linewidth}
\centering{\includegraphics[width=1\linewidth]{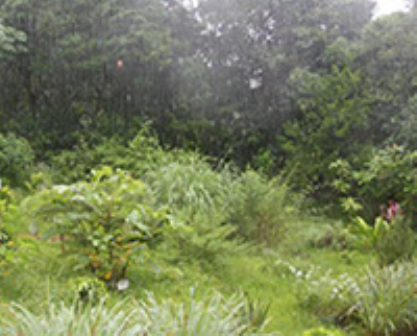}}
\end{minipage}
\hfill
\begin{minipage}{0.135\linewidth}
\centering{\includegraphics[width=1\linewidth]{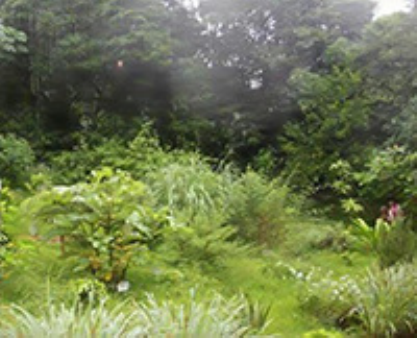}}
\end{minipage}
\hfill
\begin{minipage}{0.135\linewidth}
\centering{\includegraphics[width=1\linewidth]{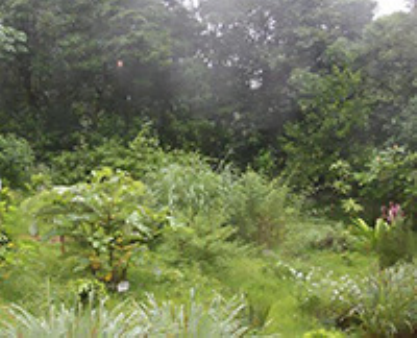}}
\end{minipage}
\hfill
\begin{minipage}{.135\linewidth}
\centering{\includegraphics[width=1\linewidth]{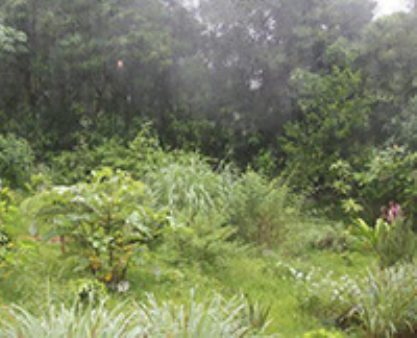}}
\end{minipage} \\ \vspace{0.5mm}
\vfill
\begin{minipage}{0.135\linewidth}
\centering{\includegraphics[width=1\linewidth]{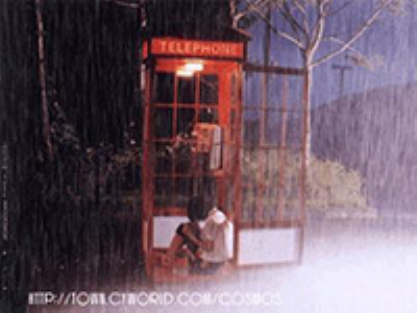}}
\end{minipage}
\hfill
\begin{minipage}{0.135\linewidth}
\centering{\includegraphics[width=1\linewidth]{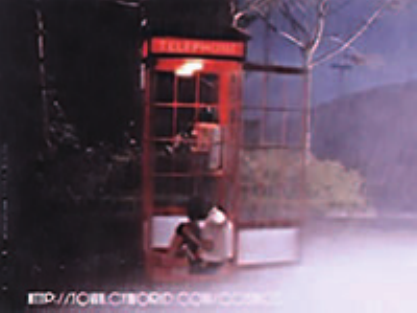}}
\end{minipage}
\hfill
\begin{minipage}{.135\linewidth}
\centering{\includegraphics[width=1\linewidth]{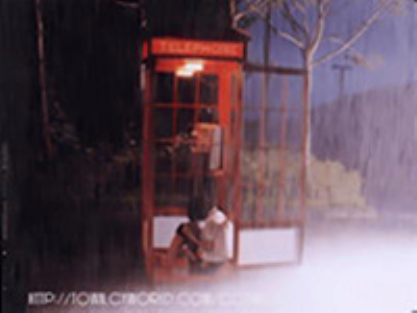}}
\end{minipage}
\hfill
\begin{minipage}{0.135\linewidth}
\centering{\includegraphics[width=1\linewidth]{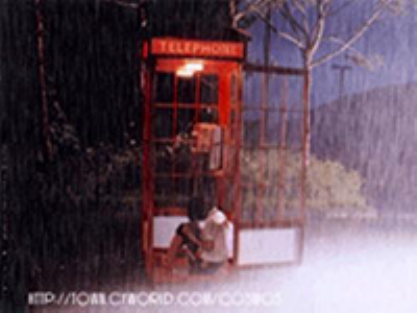}}
\end{minipage}
\hfill
\begin{minipage}{0.135\linewidth}
\centering{\includegraphics[width=1\linewidth]{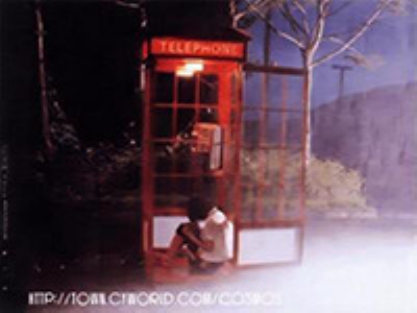}}
\end{minipage}
\hfill
\begin{minipage}{0.135\linewidth}
\centering{\includegraphics[width=1\linewidth]{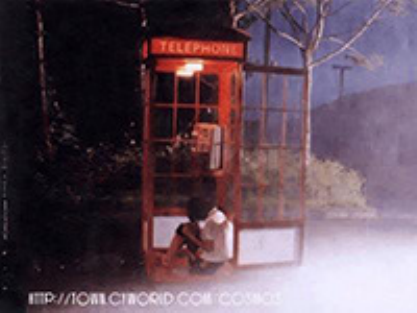}}
\end{minipage}
\hfill
\begin{minipage}{.135\linewidth}
\centering{\includegraphics[width=1\linewidth]{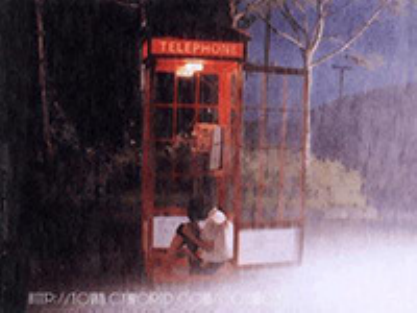}}
\end{minipage}  \\ \vspace{0.5mm}
\vfill
\begin{minipage}{0.135\linewidth}
\centering{\includegraphics[width=1\linewidth]{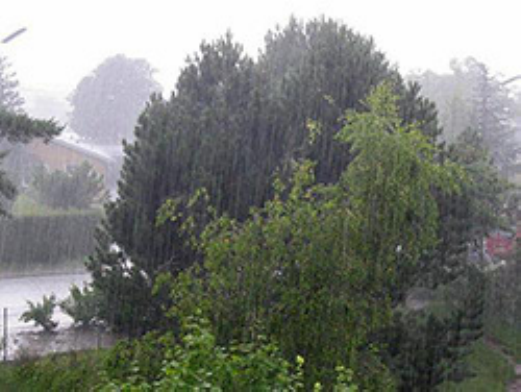}}
\end{minipage}
\hfill
\begin{minipage}{0.135\linewidth}
\centering{\includegraphics[width=1\linewidth]{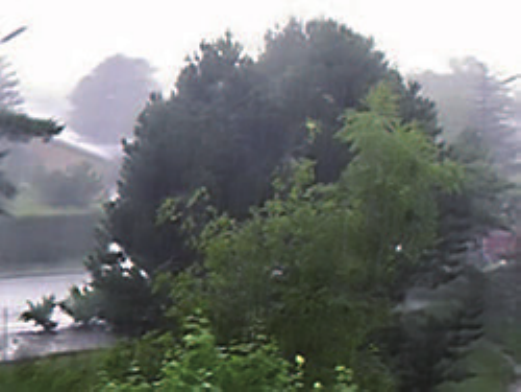}}
\end{minipage}
\hfill
\begin{minipage}{.135\linewidth}
\centering{\includegraphics[width=1\linewidth]{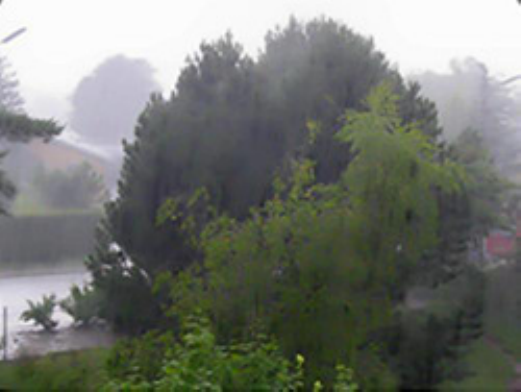}}
\end{minipage}
\hfill
\begin{minipage}{0.135\linewidth}
\centering{\includegraphics[width=1\linewidth]{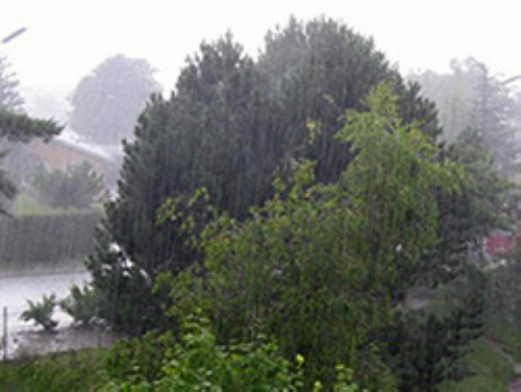}}
\end{minipage}
\hfill
\begin{minipage}{0.135\linewidth}
\centering{\includegraphics[width=1\linewidth]{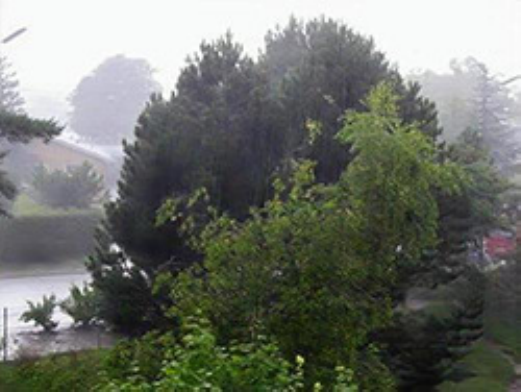}}
\end{minipage}
\hfill
\begin{minipage}{0.135\linewidth}
\centering{\includegraphics[width=1\linewidth]{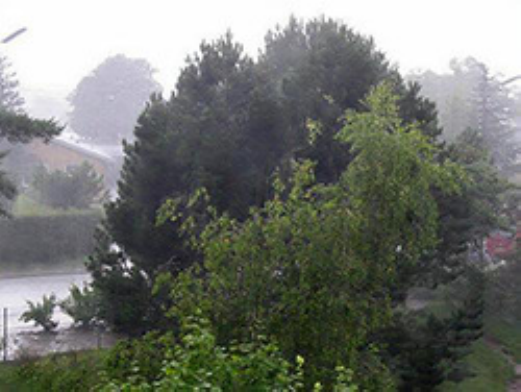}}
\end{minipage}
\hfill
\begin{minipage}{.135\linewidth}
\centering{\includegraphics[width=1\linewidth]{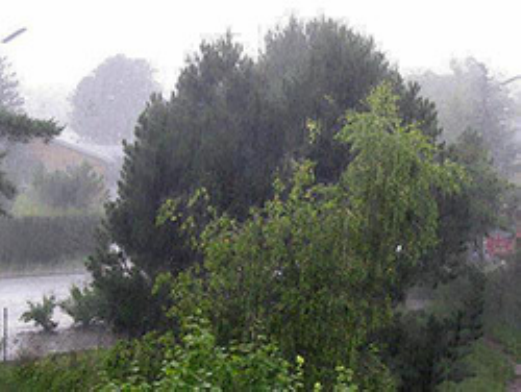}}
\end{minipage} \\ \vspace{0.5mm}
\vfill
\begin{minipage}{0.135\linewidth}
\centering{\includegraphics[width=1\linewidth]{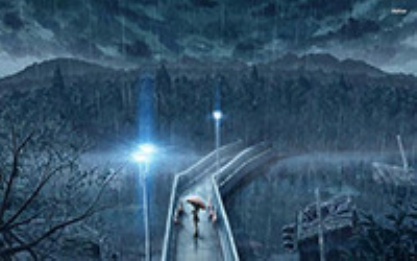}}
\end{minipage}
\hfill
\begin{minipage}{0.135\linewidth}
\centering{\includegraphics[width=1\linewidth]{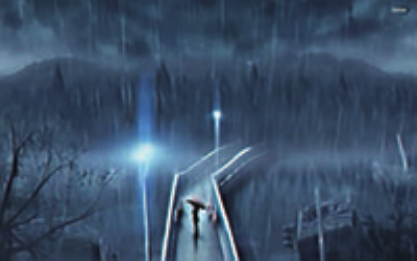}}
\end{minipage}
\hfill
\begin{minipage}{.135\linewidth}
\centering{\includegraphics[width=1\linewidth]{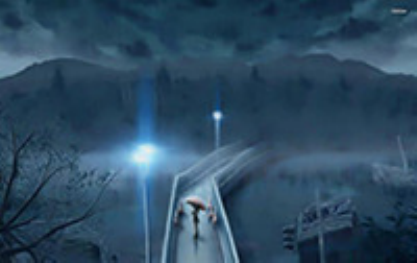}}
\end{minipage}
\hfill
\begin{minipage}{0.135\linewidth}
\centering{\includegraphics[width=1\linewidth]{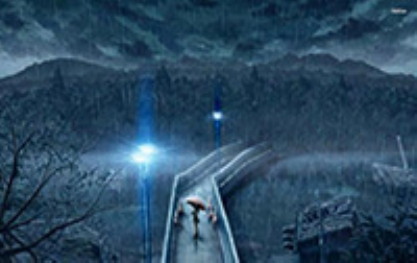}}
\end{minipage}
\hfill
\begin{minipage}{0.135\linewidth}
\centering{\includegraphics[width=1\linewidth]{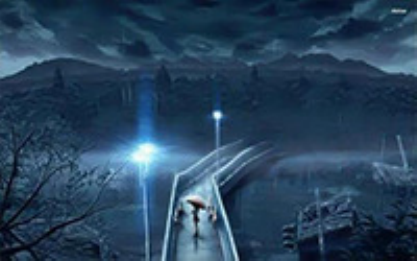}}
\end{minipage}
\hfill
\begin{minipage}{0.135\linewidth}
\centering{\includegraphics[width=1\linewidth]{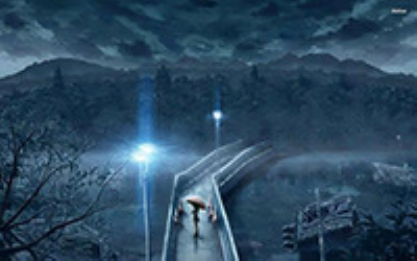}}
\end{minipage}
\hfill
\begin{minipage}{.135\linewidth}
\centering{\includegraphics[width=1\linewidth]{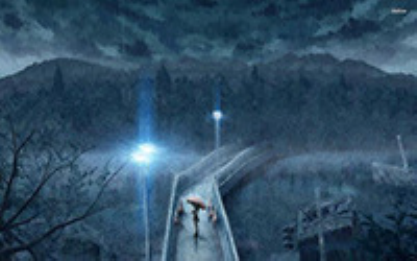}}
\end{minipage} \\ \vspace{0.5mm}
\vfill
\begin{minipage}{0.135\linewidth}
\centering{\includegraphics[width=1\linewidth]{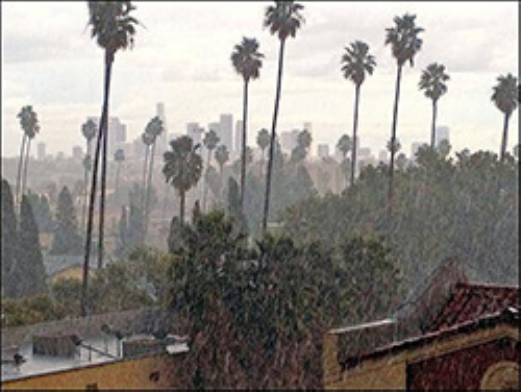}}
\centerline{(a)}
\end{minipage}
\hfill
\begin{minipage}{0.135\linewidth}
\centering{\includegraphics[width=1\linewidth]{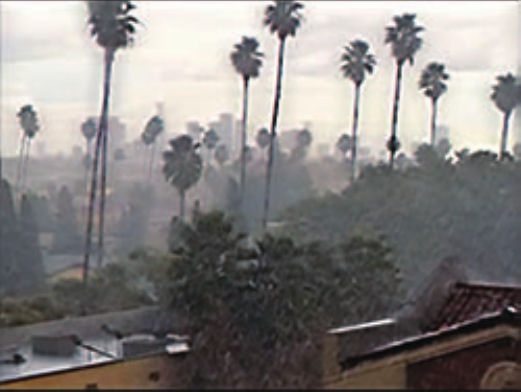}}
\centerline{(b)}
\end{minipage}
\hfill
\begin{minipage}{.135\linewidth}
\centering{\includegraphics[width=1\linewidth]{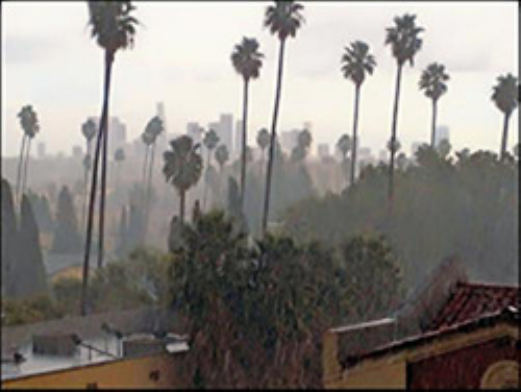}}
\centerline{(c)}
\end{minipage}
\hfill
\begin{minipage}{0.135\linewidth}
\centering{\includegraphics[width=1\linewidth]{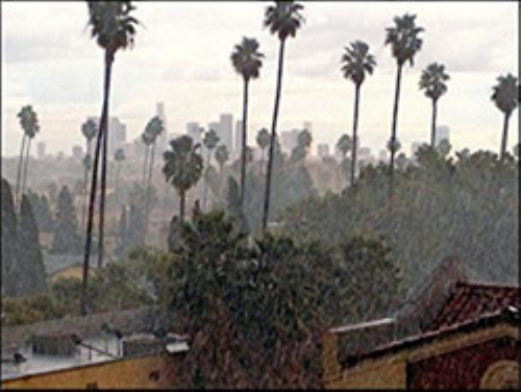}}
\centerline{(d)}
\end{minipage}
\hfill
\begin{minipage}{0.135\linewidth}
\centering{\includegraphics[width=1\linewidth]{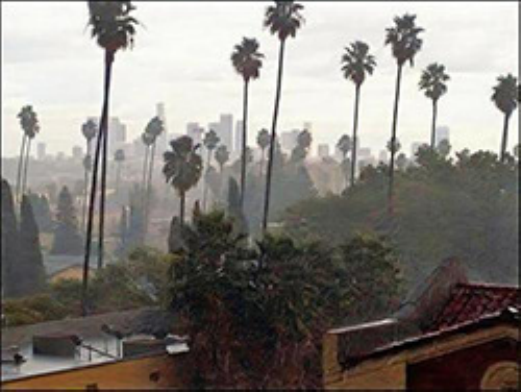}}
\centerline{(e)}
\end{minipage}
\hfill
\begin{minipage}{0.135\linewidth}
\centering{\includegraphics[width=1\linewidth]{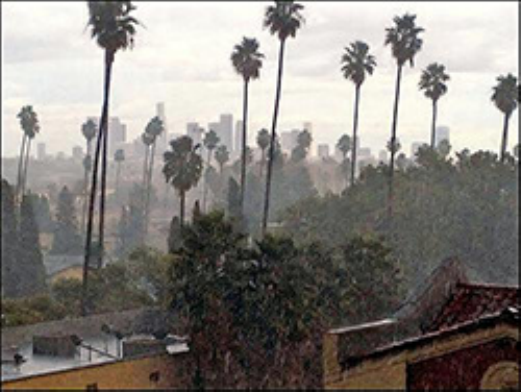}}
\centerline{(f)}
\end{minipage}
\hfill
\begin{minipage}{.135\linewidth}
\centering{\includegraphics[width=1\linewidth]{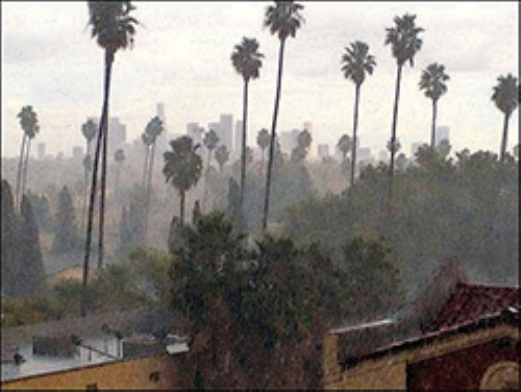}}
\centerline{(g)}
\end{minipage}
\end{center}
\caption{Some rain-removed results: (a) original light rain images, (b) results by Chen \textit{et al.} \cite{Chen_2014_CSVT}; (c) results of Li \textit{et al.} \cite{Li_2016_CVPR}, (d) results of Luo \textit{et al.} \cite{Luo_2015_ICCV}, (e) results of Zhang \textit{et al.} \cite{Zhang_2018_CVPR}, (f) results of Fu \textit{et al.} \cite{Fu_2017_CVPR} and (g) our results.}
\label{fig:visual_compare}
\end{figure*}

\begin{figure}[t]
\begin{center}
\begin{minipage}{.48\linewidth}
\centering{\includegraphics[width=1\linewidth]{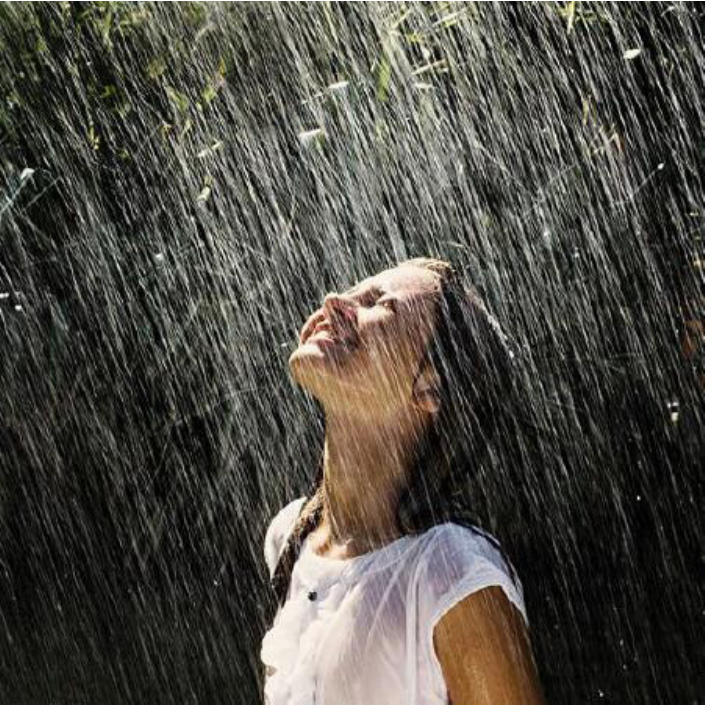}}
\centerline{(a)}
\end{minipage}
\begin{minipage}{.24\linewidth}
\centering{\includegraphics[width=1\linewidth]{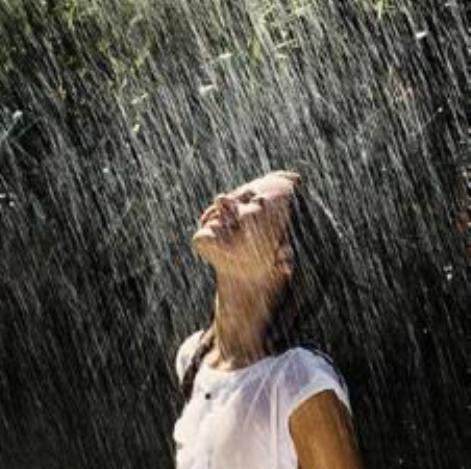}}
\centerline{(b)}
\end{minipage}
\end{center}
   \caption{(a) original heavy rain image with size $450 \times 451$; (b) the image resized to $225 \times 226$. We can see that the intensity and size of rain streaks are reduced in the resized image.}
\label{fig:rain4_resize}
\end{figure}

\subsection{Performance evaluation}

In this subsection, we evaluate the performance of our proposed rain-removal algorithm. We conduct both objective and subjective evaluations through a comparison with several state-of-the-art works. Specifically,
three very recent works which based on the traditional methods have been selected here: the work of Li \textit{et al.} in 2016 in which some patch-based priors of both the background and rain are used to separate a rain image into the rain layer and de-rained layer \cite{Li_2016_CVPR}, the work of Luo \textit{et al.} in 2015 that utilizes a nonlinear generative model to remove rain streaks from single images \cite{Luo_2015_ICCV}, and the work of Chen \textit{et al.} in 2014 that removes rain steaks by sparse coding \cite{Chen_2014_CSVT}. For deep learning based rain-removal works, we select the work \cite{Fu_2017_CVPR} by Fu \textit{et al.} to make comparisons. This work utilizes the famous ResNet which has shown its powerful ability in dealing with
various computer vision tasks.

\textbf{Complexity analysis.} \
We implement the selected methods on an Intel (R) Xeon (R) CPU E5-2643 v2 @ 3.5 GHz 3.5 GHz (2 processors) with 64G RAM, and use $256 \times 256$ color images to test the time consumption. The average time consumed by our algorithm is $7.68s$. Specifically, the detection step consumes about $5.89s$, the approximation consumes about $0.29s$, and the optimization step consumes about $1.25s$. The other small part of time is consumed by other intermediate steps. The run time consumed by the works of \cite{Chen_2014_CSVT}, \cite{Li_2016_CVPR}, \cite{Luo_2015_ICCV} are listed in the Table \ref{tab:time}. Apparently, our algorithm provides a significant speed-up (of several times to more than one degree of magnitude).

With the size change of the rain images, the consumed time is also different. In our paper, we simplify the rain imaging into a linear model and reduce the calculation complexity largely. Suppose that $N$ is the number of pixels in a given rain image $I$, $M$ is the number of candidate rain pixels after first detection, and $G$ is the number of detected rain pixels after revised by eigen color. Then the complexity for the first detection, the revision of the candidate rain pixels, and the approximation of the rain pixel are $O(N)$, $O(M)$, and $O(G)$, respectively. For the optimization of linear model, assume that $K_{i}, i=1, 2, ..., G$ are the number of rain pixels in the windows which centered at each detected rain pixel. The complexity of optimization is $O(\sum_{i=1}^{G} K_{i}) + O(G)$.

\textbf{Objective assessment.} \
In order to assess different methods quantitatively, we synthesize rain images by the screen blend model\footnote{The rendering method of rain images can be referred to \cite{Luo_2015_ICCV}.} \cite{Luo_2015_ICCV} and calculate the PSNR/SSIM \cite{Wang_2004_TIP} as the objective index.
The PSNR/SSIM values of ten synthesized rain images which are handled by traditional methods are shown in Table \ref{tab:psnrssim_rain} and several ground-truthes, corresponding rendering rain images and their rain-removed results are shown in Fig. \ref{fig:render_rain}. In order to facilitate the comparison, we list the PSNR/SSIM values at the top left of each rain-removed image in Fig. \ref{fig:render_rain}.

Overall, the method by Li \textit{et al.} \cite{Li_2016_CVPR} provides lower PSNR values. In particular, this method losses a lot useful information for images that possess many details (e.g., the second row in Fig. \ref{fig:visual_compare}). Hence, the resulted SSIM values are much lower. On the contrary, the method by Luo \textit{et al.} \cite{Luo_2015_ICCV} can remove light rain streaks and make the relatively heavy rain streaks blur. Therefore, it leads to high SSIM values. However, the PSNR values resulted by this method are still much lower. The method by Chen \textit{et al.} \cite{Chen_2014_CSVT} also losses image details, thus leading to lower SSIM values. However, the loss is not as severe as the method by Li \textit{et al.} so that its PNSR values remain to be high comparatively. It is clear that our algorithm produces better PSNR/SSIM values for a large majority of test images compared with the works which use the traditional methods. The comparisons of PSNR and SSIM with the deep learning methods \cite{Fu_2017_CVPR} and \cite{Zhang_2018_CVPR} are shown in Fig \ref{fig:psnr_ssim_graph_Fu}.
We can see from these results that our method obtains comparable PSNR values compared with \cite{Fu_2017_CVPR} and \cite{Zhang_2018_CVPR}, while the SSIM values of the deep detailed network \cite{Fu_2017_CVPR} are consistently higher than those of DID-MDN and our method for this group of images. For the deep learning methods, the involved networks are trained from thousands of rendering rain images and the corresponding groungtruthes. Hence, it is not surprising that the PSNR/SSIM values of deep learning methods are higher.

\begin{figure*}[!t]
\begin{center}
\begin{minipage}{0.115\linewidth}
\centering{\includegraphics[width=1\linewidth]{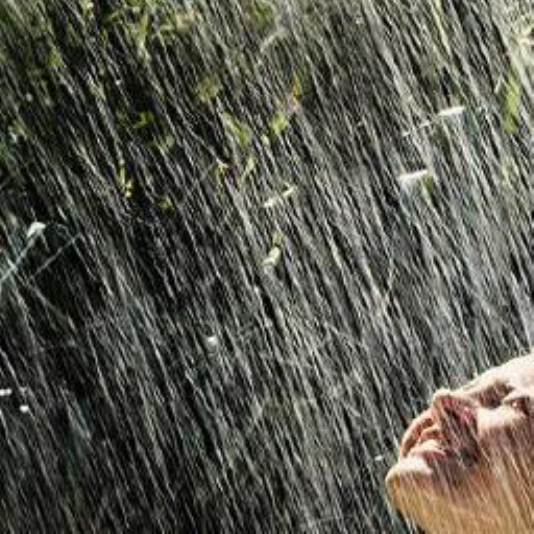}}
\end{minipage}
\hfill
\begin{minipage}{0.115\linewidth}
\centering{\includegraphics[width=1\linewidth]{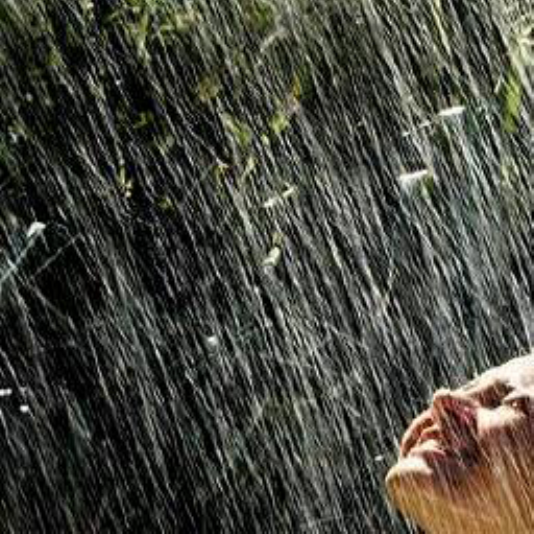}}
\end{minipage}
\hfill
\begin{minipage}{0.115\linewidth}
\centering{\includegraphics[width=1\linewidth]{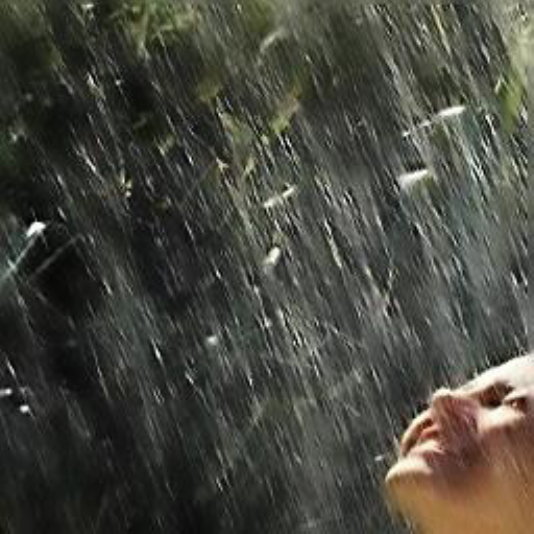}}
\end{minipage}
\hfill
\begin{minipage}{.115\linewidth}
\centering{\includegraphics[width=1\linewidth]{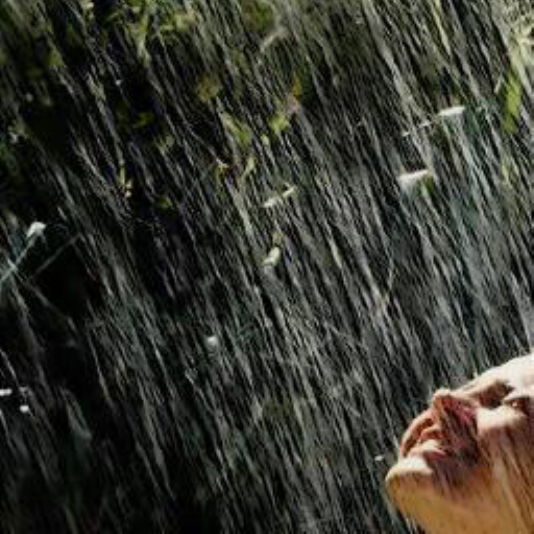}}
\end{minipage}
\hfill
\begin{minipage}{0.115\linewidth}
\centering{\includegraphics[width=1\linewidth]{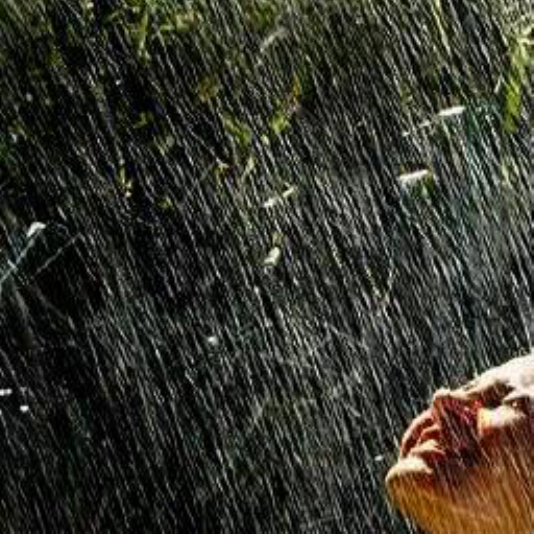}}
\end{minipage}
\hfill
\begin{minipage}{0.115\linewidth}
\centering{\includegraphics[width=1\linewidth]{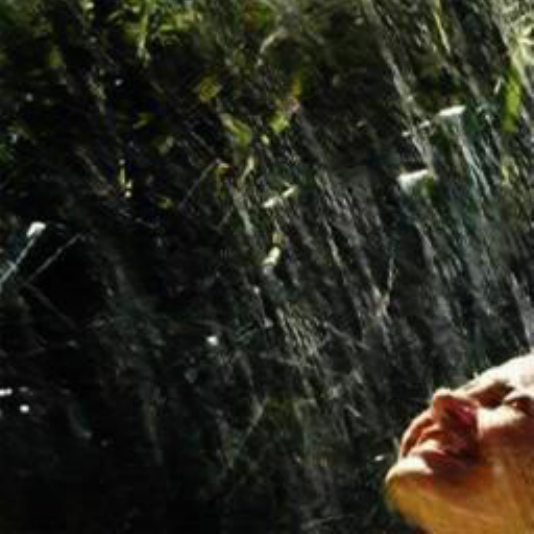}}
\end{minipage}
\hfill
\begin{minipage}{0.115\linewidth}
\centering{\includegraphics[width=1\linewidth]{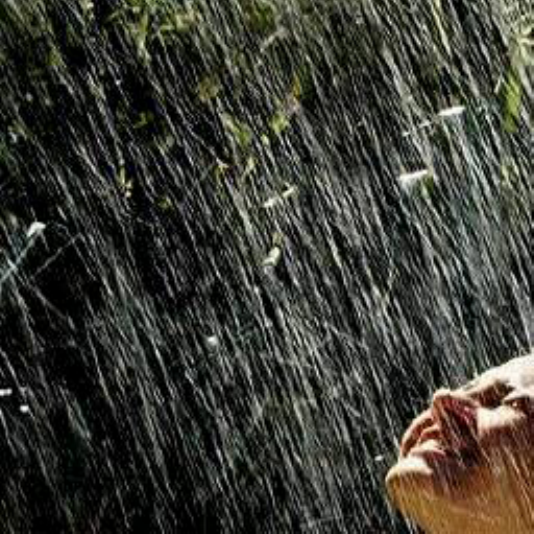}}
\end{minipage}
\hfill
\begin{minipage}{.115\linewidth}
\centering{\includegraphics[width=1\linewidth]{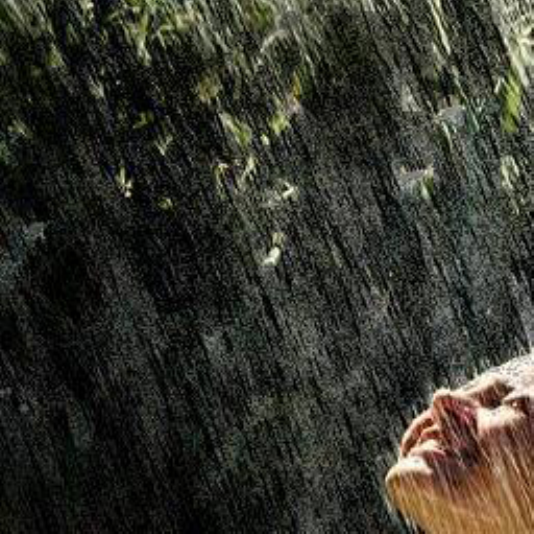}}
\end{minipage}  \\ \vspace{0.5mm}
\vfill
\begin{minipage}{0.115\linewidth}
\centering{\includegraphics[width=1\linewidth]{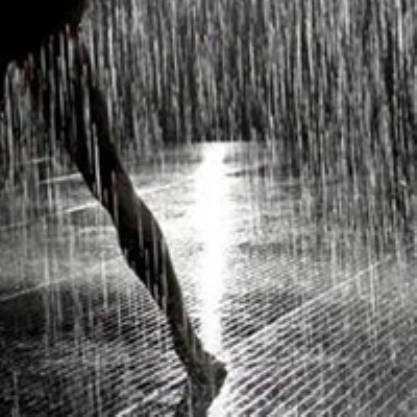}}
\end{minipage}
\hfill
\begin{minipage}{0.115\linewidth}
\centering{\includegraphics[width=1\linewidth]{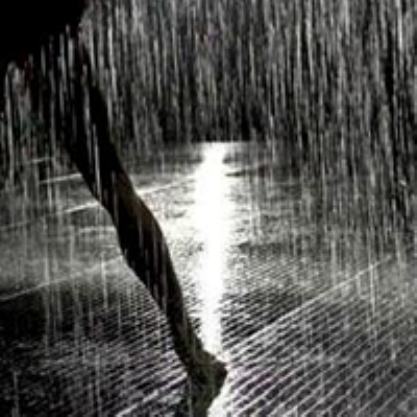}}
\end{minipage}
\hfill
\begin{minipage}{0.115\linewidth}
\centering{\includegraphics[width=1\linewidth]{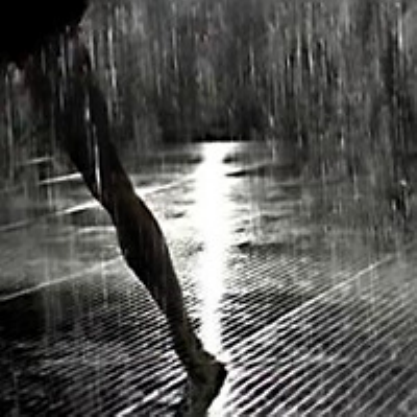}}
\end{minipage}
\hfill
\begin{minipage}{.115\linewidth}
\centering{\includegraphics[width=1\linewidth]{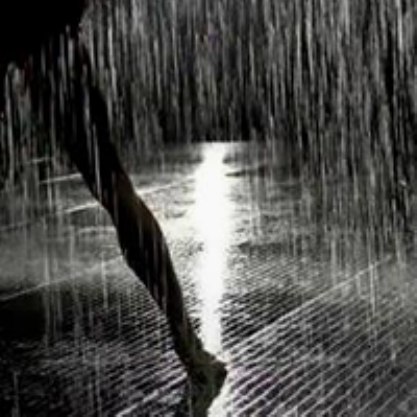}}
\end{minipage}
\hfill
\begin{minipage}{0.115\linewidth}
\centering{\includegraphics[width=1\linewidth]{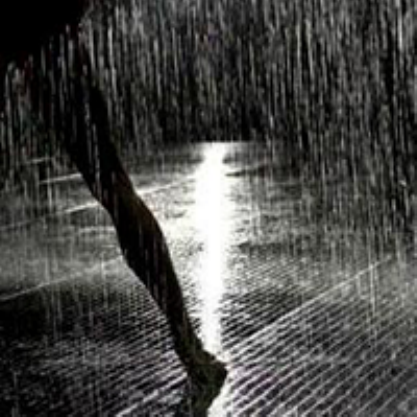}}
\end{minipage}
\hfill
\begin{minipage}{0.115\linewidth}
\centering{\includegraphics[width=1\linewidth]{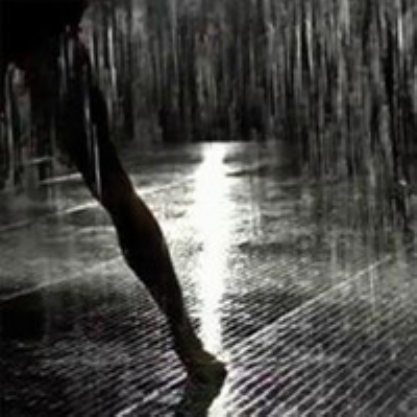}}
\end{minipage}
\hfill
\begin{minipage}{0.115\linewidth}
\centering{\includegraphics[width=1\linewidth]{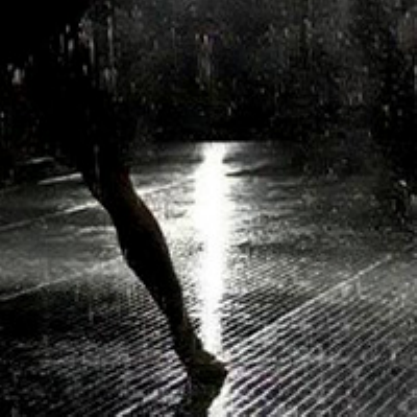}}
\end{minipage}
\hfill
\begin{minipage}{.115\linewidth}
\centering{\includegraphics[width=1\linewidth]{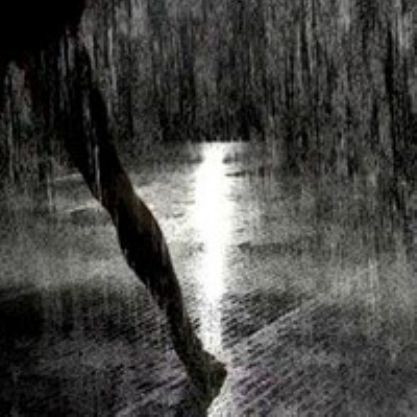}}
\end{minipage}  \\ \vspace{0.5mm}
\vfill
\begin{minipage}{0.115\linewidth}
\centering{\includegraphics[width=1\linewidth]{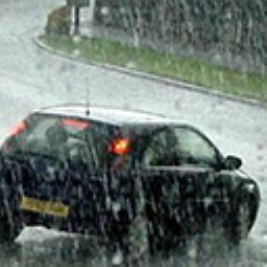}}
\centerline{(a)}
\end{minipage}
\hfill
\begin{minipage}{0.115\linewidth}
\centering{\includegraphics[width=1\linewidth]{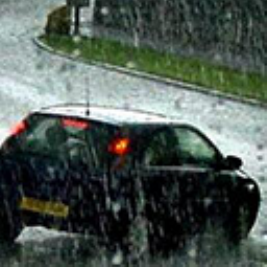}}
\centerline{(b)}
\end{minipage}
\hfill
\begin{minipage}{0.115\linewidth}
\centering{\includegraphics[width=1\linewidth]{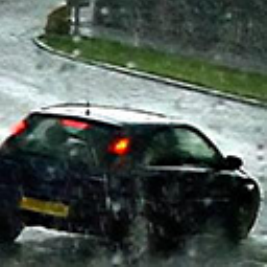}}
\centerline{(c)}
\end{minipage}
\hfill
\begin{minipage}{.115\linewidth}
\centering{\includegraphics[width=1\linewidth]{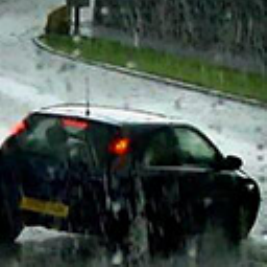}}
\centerline{(d)}
\end{minipage}
\hfill
\begin{minipage}{0.115\linewidth}
\centering{\includegraphics[width=1\linewidth]{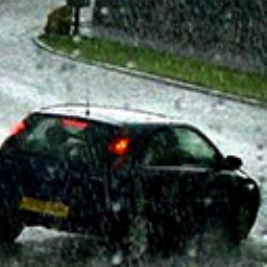}}
\centerline{(e)}
\end{minipage}
\hfill
\begin{minipage}{0.115\linewidth}
\centering{\includegraphics[width=1\linewidth]{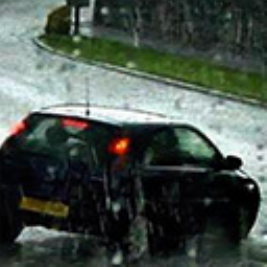}}
\centerline{(f)}
\end{minipage}
\hfill
\begin{minipage}{0.115\linewidth}
\centering{\includegraphics[width=1\linewidth]{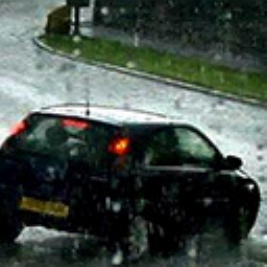}}
\centerline{(g)}
\end{minipage}
\hfill
\begin{minipage}{.115\linewidth}
\centering{\includegraphics[width=1\linewidth]{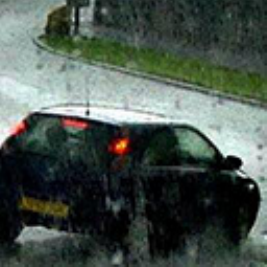}}
\centerline{(h)}
\end{minipage}
\end{center}
\caption{Some rain-removed results on heavy rain images: (a) original heavy rain images, (b) de-hazed results; (c) results by Chen \textit{et al.} \cite{Chen_2014_CSVT}; (d) results of Li \textit{et al.} \cite{Li_2016_CVPR}, (e) results of Luo \textit{et al.} \cite{Luo_2015_ICCV}, (f) results of Zhang \textit{et al.} \cite{Zhang_2018_CVPR}, (g) results of Fu \textit{et al.} \cite{Fu_2017_CVPR} and (h) our results.}
\label{fig:heavy_rain_part}
\end{figure*}

\textbf{User study.} \ To conduct a visual (subjective) evaluation on the performances of different traditional methods, 20 viewers are invited to evaluate the visual quality of different methods in terms of the following three aspects:
\begin{itemize}
\item less rain residual,
\item the maintenance of image details,
\item overall perception.
\end{itemize}

In the evaluation, 20 groups of results are selected and every group involves the results by Chen \textit{et al.} \cite{Chen_2014_CSVT}, Li \textit{et al.} \cite{Li_2016_CVPR}, Luo \textit{et al.} \cite{Luo_2015_ICCV} and our method. To ensure fairness, the results in each group are arranged randomly. For each group, the viewers are asked to select only one result which they like most. The evaluation result is shown in Table \ref{tab:statistics}. It is clear that our rain removal results are favored by a majority of viewers ($69.8\%$).

\textbf{Real rain images.} \ Taking the practical utility into consideration, we implement our algorithm on several real rain images
and compare with the selected works. The results are in Fig. \ref{fig:visual_compare}. The work by Li \textit{et al.} can remove rain streaks completely, but loss many image details, especially for the image like the second row. This is because that the patch-based priors cannot separate small image details from rain streaks well. Because the HOG descriptor used in the work by Chen \textit{et al.} cannot identify small image details either, this work is not suitable for rain images with small details (e.g., the second and third row) - edges in the rain-removed images are blurred. Besides, when the intensities of rain become higher (e.g., the fifth row), rain streaks cannot be removed well by this work. This is because that the guided filter used in this work cannot filter out all relatively bright rain streaks. For light rain streaks (e.g., the second row), the work by Luo \textit{et al.} removes rain streaks well and offers a good visual quality. Once rain streaks become relatively higher, the rain-removed performance decreases severely. Because of the detection of rain and reasonable linear model of rain imaging, the visual quality of our rain-removed results are better than selected traditional works.
We can also see from Fig. \ref{fig:visual_compare} that, for majority of light real rain images, our linear model can obtain comparable rain-removed results compared with the deep learning work \cite{Fu_2017_CVPR} and \cite{Zhang_2018_CVPR}.

\begin{figure*}[t]
\begin{center}
\begin{minipage}{.114\linewidth}
\centering{\includegraphics[width=1\linewidth]{images/test37}}
\end{minipage}
\begin{minipage}{.124\linewidth}
\centering{\includegraphics[width=1\linewidth]{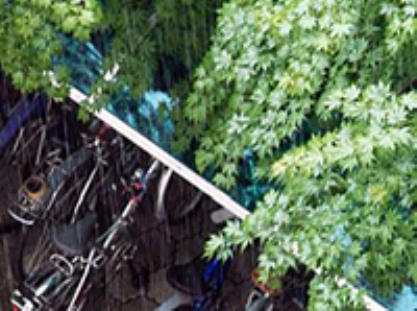}}
\end{minipage}
\begin{minipage}{.138\linewidth}
\centering{\includegraphics[width=1\linewidth]{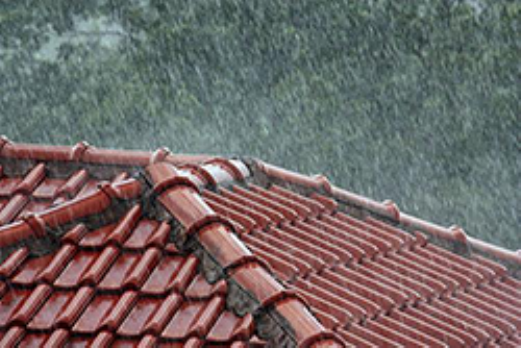}}
\end{minipage}
\begin{minipage}{.129\linewidth}
\centering{\includegraphics[width=1\linewidth]{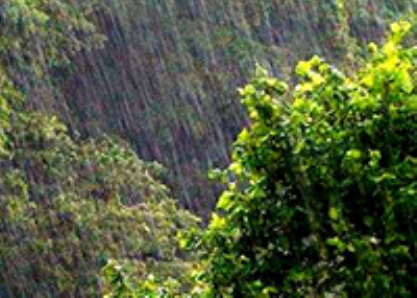}}
\end{minipage}
\begin{minipage}{.123\linewidth}
\centering{\includegraphics[width=1\linewidth]{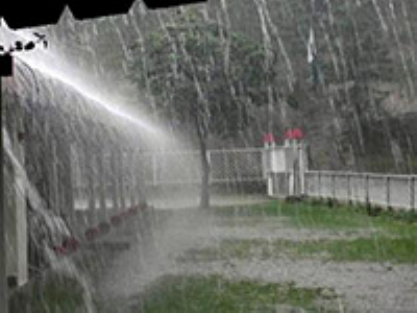}}
\end{minipage}
\begin{minipage}{.092\linewidth}
\centering{\includegraphics[width=1\linewidth]{images/rain4_dehaze_part}}
\end{minipage}
\begin{minipage}{.092\linewidth}
\centering{\includegraphics[width=1\linewidth]{images/rain5_dehaze_part}}
\end{minipage}
\begin{minipage}{.138\linewidth}
\centering{\includegraphics[width=1\linewidth]{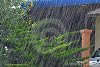}}
\end{minipage} \\
\vspace{0.5mm}
\begin{minipage}{.114\linewidth}
\centering{\includegraphics[width=1\linewidth]{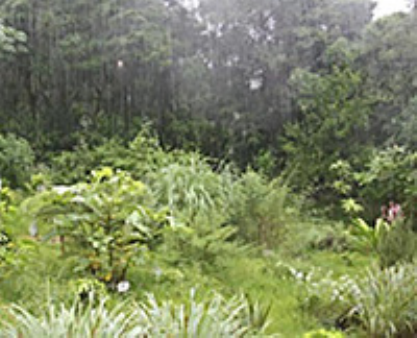}}
\end{minipage}
\begin{minipage}{.124\linewidth}
\centering{\includegraphics[width=1\linewidth]{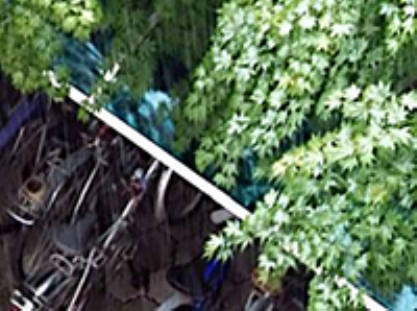}}
\end{minipage}
\begin{minipage}{.138\linewidth}
\centering{\includegraphics[width=1\linewidth]{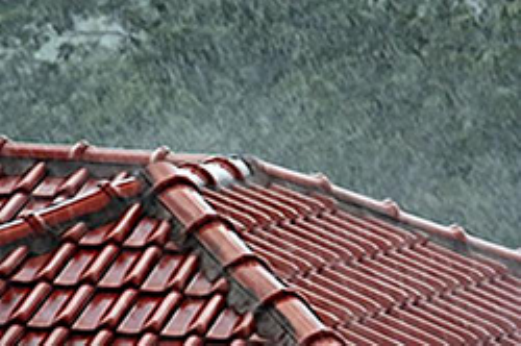}}
\end{minipage}
\begin{minipage}{.129\linewidth}
\centering{\includegraphics[width=1\linewidth]{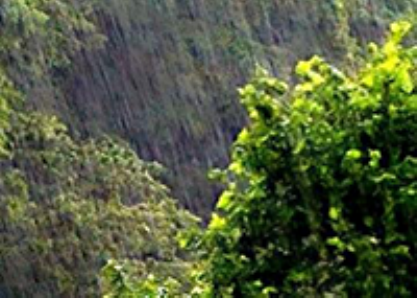}}
\end{minipage}
\begin{minipage}{.123\linewidth}
\centering{\includegraphics[width=1\linewidth]{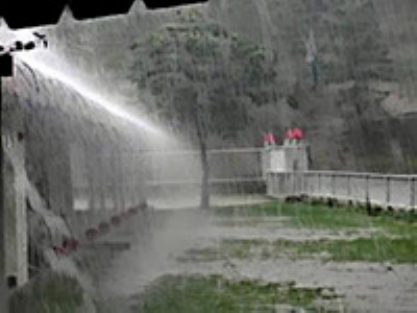}}
\end{minipage}
\begin{minipage}{.092\linewidth}
\centering{\includegraphics[width=1\linewidth]{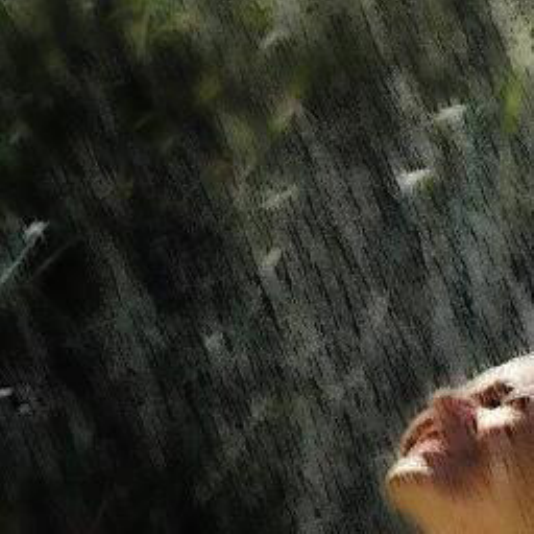}}
\end{minipage}
\begin{minipage}{.092\linewidth}
\centering{\includegraphics[width=1\linewidth]{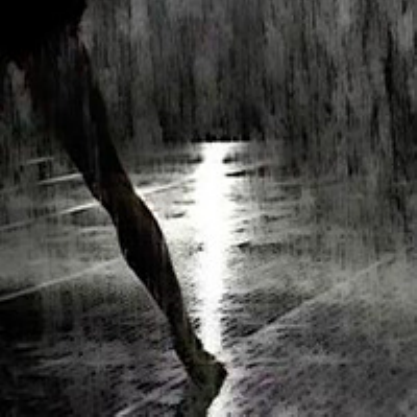}}
\end{minipage}
\begin{minipage}{.138\linewidth}
\centering{\includegraphics[width=1\linewidth]{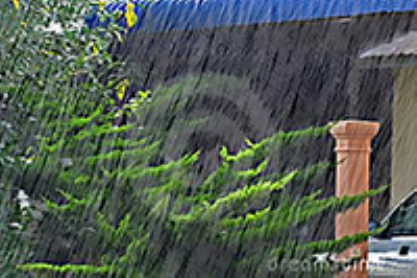}}
\end{minipage} \\
\vspace{0.5mm}
\begin{minipage}{.114\linewidth}
\centering{\includegraphics[width=1\linewidth]{images/test37_g}}
\end{minipage}
\begin{minipage}{.124\linewidth}
\centering{\includegraphics[width=1\linewidth]{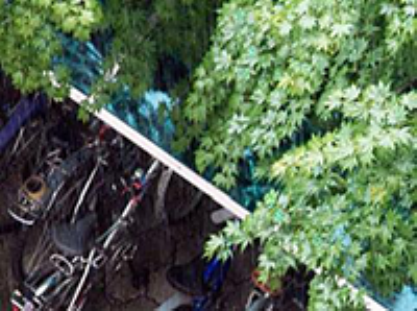}}
\end{minipage}
\begin{minipage}{.138\linewidth}
\centering{\includegraphics[width=1\linewidth]{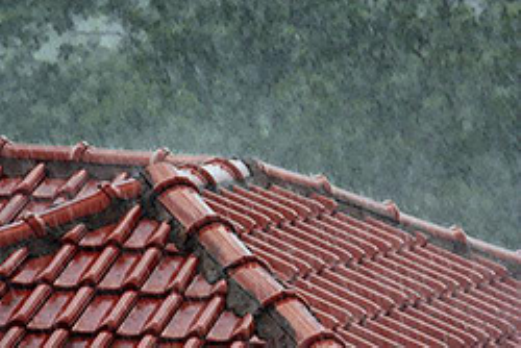}}
\end{minipage}
\begin{minipage}{.129\linewidth}
\centering{\includegraphics[width=1\linewidth]{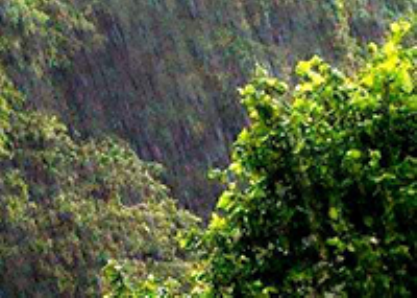}}
\end{minipage}
\begin{minipage}{.123\linewidth}
\centering{\includegraphics[width=1\linewidth]{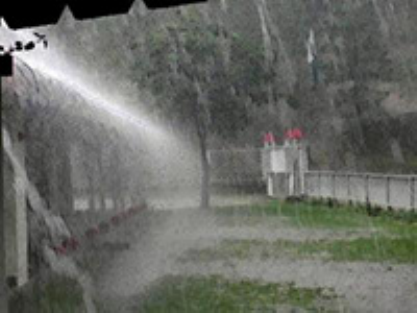}}
\end{minipage}
\begin{minipage}{.092\linewidth}
\centering{\includegraphics[width=1\linewidth]{images/rain4_dehaze_part_derain}}
\end{minipage}
\begin{minipage}{.092\linewidth}
\centering{\includegraphics[width=1\linewidth]{images/rain5_dehaze_part_derain}}
\end{minipage}
\begin{minipage}{.138\linewidth}
\centering{\includegraphics[width=1\linewidth]{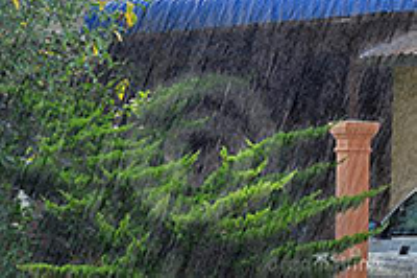}}
\end{minipage}
\end{center}
   \caption{From top to down: rain images, results by the hierarchical approach, results by the linear model.}
\label{fig:compare_with_TIP}
\end{figure*}

\begin{figure}[t]
\begin{center}
\begin{minipage}{.48\linewidth}
\centering{\includegraphics[width=1\linewidth]{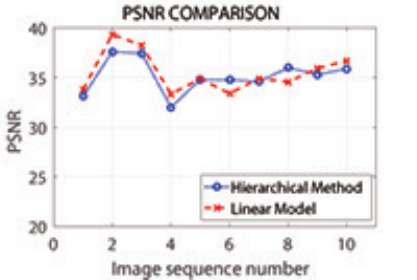}}
\centerline{(a)}
\end{minipage}
\begin{minipage}{.48\linewidth}
\centering{\includegraphics[width=1\linewidth]{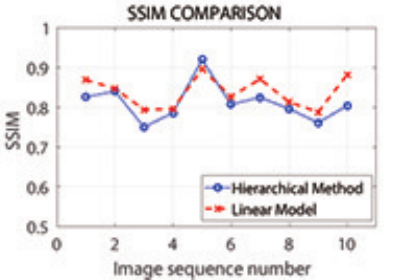}}
\centerline{(b)}
\end{minipage}
\end{center}
   \caption{(a) The PSNR values of two methods. (b) The SSIM values of two methods.}
\label{fig:psnr_ssim_graph}
\end{figure}

\subsection{Challenging from heavy rain images}
In order to further verify our linear model, we consider the challenging from heavy rain images. In this scenario, fog often appears in the image so that we propose to implement a de-haze preprocessing before removing rain streaks.

There are several excellent algorithms for the de-hazing task. In \cite{Ren_2016_ECCV}, the authors learnt the mapping between haze images and their corresponding transmission maps by a multi-scale deep neutral network, and the resulting performance outperforms many previous
de-haze works (e.g. the dark channel method \cite{He_2011_PAMI} by He \textit{et al.}). Hence, \cite{Ren_2016_ECCV} is selected in our work.

It is necessary to point out that the rain-removing methods presented in \cite{Chen_2014_CSVT} and \cite{Luo_2015_ICCV} are very memory-consuming, which may lead to memory overflow even on a highly-configured computer. Hence, a down-sizing has been included in \cite{Chen_2014_CSVT} and \cite{Luo_2015_ICCV} before the rain-removing, while the output image is up-sized back. One example is shown in Fig. \ref{fig:rain4_resize}, from which we can see that the intensity and size of rain streaks are reduced. This would make the removal of rain streaks much easier.

In order to make the performance comparison among different algorithms fair and accurate, we need to implement them on the same rain image. To this end, we crop a $256\times 256$ patch from each original rain image. Then, the rain-removing methods in \cite{Chen_2014_CSVT} and \cite{Luo_2015_ICCV} can be implemented without resizing.

Some comparison results are presented in Fig. \ref{fig:heavy_rain_part}. Several observations can be made from these results. (1) The method of \cite{Li_2016_CVPR} can not deal with heavy rain images well and the performance of \cite{Luo_2015_ICCV} is also sensitive to the scale of rain streaks. (2) The method of \cite{Chen_2014_CSVT} removes majority of rain streaks and produce better results as compared to the above two methods. (3) Overall, our linear model produces the best results, no matter in image-detail preservation or rain-streak removal compared with the selected works which are based
on traditional methods. Besides, our linear model has a very low computational complexity and is not memory-consuming. (4) The work by Fu \textit{et al.} can not remove very heavy rain in the image of the first line. The work by Zhang \textit{et al.} has done a good job in removing
heavy rain streaks. Our method also removes heavy rain streaks and makes the image details clearer. For the other two images, these three methods obtain comparable rain-removed results. Although deep learning methods have made great differences in many fields, they still cannot obtain very good results for some images whose patterns are not included in the training set. One advantage of the traditional methods is that they usually produce relatively stable results.

\subsection{Comparisons with our previous hierarchical approach}

In \cite{Wang_2017_TIP}, we developed a rain/snow removal algorithm based on a hierarchical approach and it also needs to detect rain/snow. We make some comparisons between these two works in this subsection.

In this work, we use the rain/snow detection method in \cite{Wang_2017_TIP} to detect rain streaks initially. As we mentioned in \cite{Wang_2017_TIP},
this is an over-detection method and some non-rain details will be mis-detected as rain streaks. In \cite{Wang_2017_TIP}, the lost information
is complemented by a 3-layer extraction of non-rain/snow details. In this work, we improve rain's detection by eigen color property, which has revised many mis-detected non-rain details.

More specifically, in \cite{Wang_2017_TIP}, an over-complete dictionary is learnt first and we identify rain/snow dictionary atoms by the characteristics of rain/snow. Rain/snow and non-rain/snow components are reconstructed by sparse coding, and the non-rain/snow component forms the first layer of the non-rain/snow details. Then, by combining detection of rain/snow and guided filter, we extract the second layer of non-rain/snow details from the rain/snow component that is obtained by the sparse reconstruction. Finally, in order to enhance image contrast, the third layer of non-rain/snow details is extracted by the developed sensitivity of variance across color channels (SVCC) descriptor. After adding the low frequency and three layers of image details together, we obtain the final rain/snow-removed results. In this work, we derive a linear imaging model of rain and remove rain by training our linear model.

Because of the dictionary learning method used in the hierarchical approach \cite{Wang_2017_TIP}, its time consumption is much bigger than the linear model. When tested on $256 \times 256$ images, the average time consumed by the hierarchical approach is $83.38$ seconds, while our linear model only needs $7.68$ seconds as mentioned earlier. In Fig. \ref{fig:compare_with_TIP}, we show some rain removal results of these two methods. For light to medium rain streaks (the first five images), these two methods produce comparable results. However, for rain images that contains many small image details (e.g., the first and fourth ones), the linear model maintains more image details and keeps better structural similarity with original images. When encounterring heavy rain images (the last three images), the hierarchical approach removes more rain streaks, but the linear model method can preserve more image details.

Some objective comparisons are evaluated in terms of PSNR/SSIM for these two methods, as listed in Fig. \ref{fig:psnr_ssim_graph}. It can be seen that the linear model obtains lightly better PSNR/SSIM values for most images. This is mainly due to that our linear model will not change the non-rain areas of an
image and the linear model describes the imaging of rain more accurately.

\subsection{Limitations and future works}

Our linear model has shown good rain removal performances in several important aspects and possesses good robustness no matter for common rain images or for challenging heavy rain images. Like most of algorithms, our method still have its own limitations. Rain streaks are over-detected in our algorithm, a few of non-rain details which have similar intensity and color to rain streaks can be mis-regarded as rain. Though these mis-detections are only a small part, the performance of our algorithm will become better if we can make the detection more accurate. In our future work, we will try our best to improve the detection of rain streaks by deep learning methods. Besides, we will try to combine our linear model with deep learning to solve the heavy rain tasks. We believe this will make a great difference if it can be realized.

\section{Conclusions}
\label{sec:conclusion}
In this paper, we derive a simple linear model $p=\alpha s + \beta$ to describe the physical principle of imaging rain pixels. In order to remove rain streaks in a rain image, we first detect rain streaks by two characteristics of rain streaks. Once the binary location map of rain pixels is obtained, the original intensity of each rain pixel is
approximated by a weighted average of all neighboring non-rain pixels.
For every rain pixel, we train the parameters involved in the linear model. Once the parameters are determined for a rain pixel, its rain-removed intensity can be calculated by plugging the observed intensity of the rain pixel into the model. Subjective and objective evaluations demonstrate that our algorithm outperforms several state-of-the-art traditional methods for rain-removal. Compared with deep learning based method, our linear model can obtain comparable rain-removed results no matter for light rain images or for majority of heavy rain images. For some very heavy rain images, our model even outperforms the deep learning based algorithms. Moreover, our algorithm offers a significant speed-up of several times to more than one degree of magnitude compared with the selected traditional methods.

\ifCLASSOPTIONcaptionsoff
  \newpage
\fi

\end{document}